\documentclass[acmtog]{acmart}

\usepackage{multirow}
\usepackage{colortbl}
\usepackage{xspace}
\usepackage{comment}
\usepackage{stfloats}

\graphicspath{{figures/}}



\makeatletter
\@ifundefined{XR@test}{}{%
  \let\XR@test@orig\XR@test
  \long\def\XR@test#1#2#3#4\XR@{%
    \def\XR@tmplabel{#2}%
    \def\XR@totpages{TotPages}%
    \in@{tocindent}{#2}%
    \ifin@
      \ifeof\@inputcheck\expandafter\XR@aux\else\expandafter\XR@read\fi
    \else\ifx\XR@tmplabel\XR@totpages
      \ifeof\@inputcheck\expandafter\XR@aux\else\expandafter\XR@read\fi
    \else
      \XR@test@orig{#1}{#2}{#3}#4\XR@
    \fi\fi
  }%
}
\makeatother

\definecolor{suppColor}{rgb}{0.659, 0, 0.652}
\newcommand{\supp}[1]{\textcolor{suppColor}{#1}}

\newcommand{\vid}{\ensuremath{V}\xspace}
\newcommand{\hdrvid}{\ensuremath{H}\xspace}
\newcommand{\sdrvid}{\ensuremath{V}\xspace}
\newcommand{\latentsdrvid}{\ensuremath{\tilde{V}}\xspace}
\newcommand{\queries}{\ensuremath{Q}\xspace}
\newcommand{\keys}{\ensuremath{K}\xspace}

\newcommand{\noise}{\boldsymbol{\epsilon}}

\newcommand{\radiance}{\ensuremath{R}\xspace}

\newcommand{\weight}{\ensuremath{W}\xspace}

\newcommand{\numframes}{\ensuremath{F}\xspace}

\newcommand{\cond}{\ensuremath{\mathcal{C}}\xspace}

\makeatletter
\newcommand{\suppmaketitle}[1]{%
  \begingroup
  \def\@title{#1}%
  \let\@subtitle\@empty
  \let\@translatedtitle\@empty
  \let\@translatedsubtitle\@empty
  \@mktitle
  \twocolumn[\box\mktitle@bx]%
  \endgroup
}
\makeatother
\copyrightyear{2026}
\acmYear{2026}
\setcopyright{cc}
\setcctype{by}
\acmConference[SA Conference Papers '26]{SIGGRAPH Asia 2026 Conference Papers}{December 01--04, 2026}{Kuala Lumpur, Malaysia}
\acmBooktitle{SIGGRAPH Asia 2026 Conference Papers (SA Conference Papers '26), December 01--04, 2026, Kuala Lumpur, Malaysia}
\acmDOI{10.1145/3829340.3842206}
\acmISBN{979-8-4007-2842-6/2026/12}

\definecolor{tabfirst}{rgb}{1, 0.7, 0.7}
\definecolor{tabsecond}{rgb}{1, 0.85, 0.7}
\definecolor{tabthird}{rgb}{1, 1, 0.7}

\newcommand{\eref}[1]{Eq.~\eqref{#1}}

\newcounter{todos}
\AtEndDocument{\ifnum\value{todos}>0 \PackageWarning{TODOS}{There are \arabic{todos} todos left in this paper! Fix them before submitting the paper!} \fi}

\makeatletter
\DeclareRobustCommand\onedot{\futurelet\@let@token\@onedot}
\def\@onedot{\ifx\@let@token.\else.\null\fi\xspace}

\def\ie{{\it i.e}\onedot}

\def\etal{{\it et al}\onedot}
\makeatother

\begin{document}

\title{Generating HDR Video from SDR Video}

\author{SaiKiran Tedla}
\orcid{0000-0002-2679-2881}
\email{tedlasai@yorku.ca}
\affiliation{%
	\institution{Sony AI}
	\city{Tokyo}
	\country{Japan}
}
\affiliation{%
	\institution{York University}
	\city{Toronto}
	\country{Canada}
}

\author{Francesco Banterle}
\orcid{0000-0002-6374-6657}
\email{francesco.banterle@isti.cnr.it}
\affiliation{%
	\institution{ISTI-CNR}
	\city{Pisa}
	\country{Italy}
}

\author{Trevor Canham}
\orcid{0000-0003-4830-4786}
\email{tcanham@yorku.ca}
\affiliation{%
	\institution{York University}
	\city{Toronto}
	\country{Canada}
}

\author{Karanpreet Raja}
\orcid{0009-0009-1559-6883}
\email{kraja@yorku.ca}
\affiliation{%
	\institution{York University}
	\city{Toronto}
	\country{Canada}
}

\author{David B. Lindell}
\orcid{0000-0002-6999-3958}
\email{lindell@cs.toronto.edu}
\affiliation{%
	\institution{University of Toronto}
	\city{Toronto}
	\country{Canada}
}

\author{Kiriakos N. Kutulakos}
\email{kyros@cs.toronto.edu}
\orcid{0000-0002-5165-902X}
\affiliation{%
	\institution{University of Toronto}
	\city{Toronto}
	\country{Canada}
}

\author{Jiacheng Li}
\orcid{0000-0002-4215-6754}
\email{jiacheng.li@sony.com}
\affiliation{%
	\institution{Sony AI}
	\city{Tokyo}
	\country{Japan}
}

\author{Feiran Li}
\orcid{0000-0002-2585-506X}
\email{feiran.li@sony.com}
\affiliation{%
	\institution{Sony AI}
	\city{Tokyo}
	\country{Japan}
}

\author{Daisuke Iso}
\orcid{0009-0006-7083-9735}
\email{daisuke.iso@sony.com}
\affiliation{%
	\institution{Sony AI}
	\city{Tokyo}
	\country{Japan}
}

\begin{teaserfigure}
  \centering
  \includegraphics[width=\textwidth]{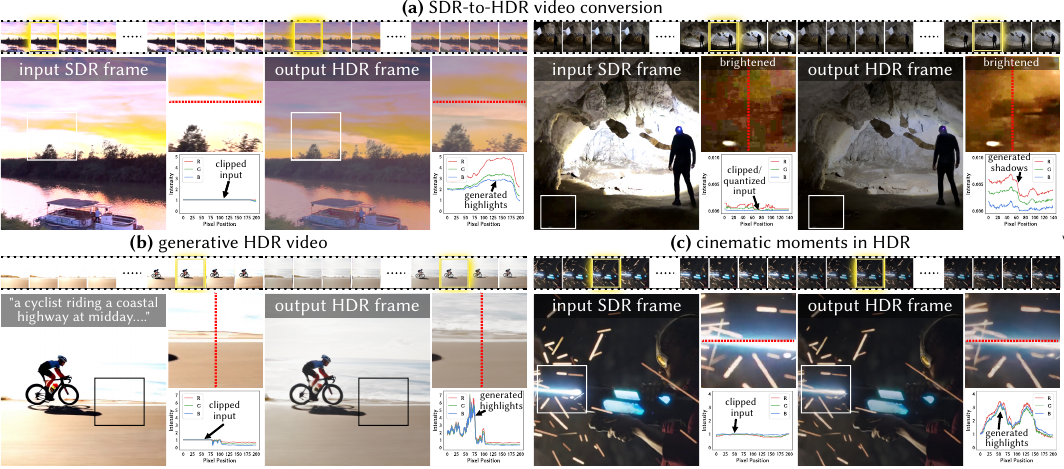}
\caption{Our method lifts casual SDR video to temporally consistent HDR by harnessing large-scale generative video models. For each example, the film strip records the full temporal sequence (highlighted frame shown below), and intensity line scans (along the red dashed line) compare the RGB profiles of the SDR input and HDR output. Scanline values show relative, unitless radiance, with the SDR input anchored to $[0,1]$; our HDR output extends this scale (up to $16{\times}$) by recovering detail lost to highlight clipping and shadow quantization. \textbf{(a)}~SDR-to-HDR video conversion. \textit{Left}: a sunset river scene whose sky is fully clipped in SDR; our model synthesizes photorealistic warm-sky highlight detail. \textit{Right}: an under-lit cave interior where the SDR signal is quantized, and the human silhouette is lost against the background; our model recovers wall texture and figure-ground separation from the quantized shadows. \textbf{(b)}~Fully generative HDR: an SDR clip synthesized from the text prompt \emph{``a cyclist riding a coastal highway at midday...''} is lifted to HDR by our model, producing a coastal scene with separated sky highlights and foreground shadow detail. \textbf{(c)}~Cinematic moments: a canonical SDR frame from \textit{StEM2} (\copyright~American Society of Cinematographers, 2022) is re-graded to HDR, simultaneously recovering energy-beam highlights and arm shadow detail. Please see the \supp{supplementary webpage} for videos.}
\label{fig:teaser}
\end{teaserfigure}

\begin{abstract}
The high dynamic range (HDR) video ecosystem is approaching maturity, but the problem of upconverting legacy standard dynamic range (SDR) videos persists without a convincing solution.
We propose a framework for HDR video synthesis from casual SDR footage by leveraging a large-scale generative video model. We introduce a Multi-Exposure Video Model (MEVM) that can predict exposure-bracketed linear SDR video sequences from a single nonlinear SDR video input.
We further propose a learnable Video Merging Model (VMM) that merges the predicted exposure-bracketed video into a high-quality HDR sequence while preserving detail in both shadows and highlights. 
Extensive experiments, quantitative and qualitative evaluation, and a user study demonstrate that our approach enables robust HDR conversion for in-the-wild examples from casual consumer videos and even cinematic footage. Finally, our model can support HDR synthesis pipelines built upon existing SDR generative video models. The project webpage is at \href{https://sdr2hdrvideo.github.io}{\texttt{sdr2hdrvideo.github.io}}
\end{abstract}
\fancypagestyle{firstpagestyle}{%
  \fancyhf{}%
  \fancyhead[R]{\footnotesize\thepage}%
  \renewcommand{\headrulewidth}{0pt}%
}
\maketitle
\fancyhf{}
\fancyhead[R]{\footnotesize\thepage}
\renewcommand{\headrulewidth}{0pt}
\flushbottom

\section{Introduction}
\label{sec:intro}

The last decade has seen major developments in the capture, post-processing, encoding, transmission, and display of High Dynamic Range (HDR) images and videos~\cite{miller2013perceptual}.
HDR displays are now standard on consumer devices such as smartphones and laptops~\cite{Banterle:2017}. These displays offer immersive experiences by approaching contrast ratios of real-world scenes and emitting luminance from 0.01 to 1000 cd/$m^2$ (nits)~\cite{reinhard20}. 
Moreover, by leveraging developments in capture and post-production, the television and cinema industries already produce a large amount of content intended for HDR display \cite{quero2025hdr10netflix}.

Against this backdrop of rapid HDR adoption, the upconversion of existing SDR video poses a major challenge: SDR videos have been produced for many decades; they often contain clipped highlight details and crushed shadows; and have been encoded and compressed with SDR reproduction in mind.
Naïve upconversion of such videos to HDR can amplify the appearance of such artifacts --- clipped highlights appearing very bright, shadowed regions lacking detail, intensity quantization becoming much more apparent --- resulting in a low-quality visual experience.
Despite considerable efforts to overcome these issues using classical methods \cite{akyuz2007hdr, didyk2008enhancement}, CNNs \cite{eilertsen2017hdr, marnerides2018expandnet, Santos20SIHDR, endo2017deep}, transformers \cite{wang2025aim}, GANs \cite{wang2023glowgan}, and image diffusion models~\cite{wang2025lediff},  convincing results for this problem remain elusive, particularly in the context of video. 

In this work, we revisit the problem of generating HDR video from casual SDR footage by harnessing the power of a large video generation model. 
Video models are trained on billions of images and millions of videos and thus provide very strong priors about dynamic appearance.
Repurposing an off-the-shelf video model for HDR upconversion is 
non-trivial because these models are trained almost exclusively on compressed, SDR-display-encoded content with 8-bit intensities.
As a result, the autoencoders these models are built on are poorly suited to represent HDR video frames (Figure \ref{fig:vae_issues}) and retraining them for HDR video generation is currently infeasible due to the lack of large-scale HDR video data~\cite{froehlich14, wang2025lediff}.

To overcome these challenges, we drew inspiration from one of the earliest photography techniques: exposure bracketing \cite{gray_1856,adams1983examples, debevec1997hdr}. Our key insight is that the task of \textit{generating an exposure-bracketed image sequence is inherently compatible with the latent space of pre-trained SDR video models because each image in the sequence is an SDR image.} 
Thus, instead of attempting to generate HDR videos directly, we finetune the video model to generate a short bracket sequence for every frame of an SDR video.
The output of the model can be thought of as $N$ videos recorded simultaneously by $N$ synchronized cameras with different exposure settings.
With these videos on hand, HDR upconversion reduces to the classical problem of converting a bracketed sequence of $N$ SDR images into a single HDR frame.
We show that this two-stage approach is effective in inpainting under-exposed and over-exposed regions of the SDR source video and enables robust HDR video upconversion without specialized HDR capture protocols or large-scale HDR video training datasets. 
The main contributions of this work are summarized as follows: 

\begin{itemize}
\item We present, to our knowledge, the first bracketed-generation approach for SDR-to-HDR conversion of casual videos. Our framework can also be seamlessly combined with off-the-shelf text or image-conditioned SDR video generation models to form a complete HDR generation pipeline.
\item We propose a Multi-Exposure Video Model (MEVM) that leverages large-scale SDR generative video pretraining to predict linear exposure-bracketed SDR videos, and a learnable Video Merging Model (VMM) that merges generated exposure-bracketed SDR videos into HDR videos.
\item Extensive experiments demonstrate that our method achieves state-of-the-art perceptual quality for SDR-to-HDR video conversion and HDR video content generation.
\end{itemize}

\section{Related Work}
\label{sec:relatedworks}

We organize prior work by the capture protocol it requires, then turn to the generative video models we build on, and finally to the most relevant SDR-to-HDR conversion literature.

\subsection{HDR Capture Techniques}

Most HDR pipelines obtain dynamic range through \emph{capture-time exposure coding}---deliberate exposure variation across time, space, or both. The classical approach captures multiple LDR frames at different shutter times and merges them into scene-referred linear radiance~\cite{Mann1995ONB,debevec1997hdr}. For dynamic scenes, a substantial line of work suppresses ghosting via patch matching and optical flow~\cite{sen2012robust,kalantari2017deep}, UNets~\cite{wu2018deep}, attention or transformer-based alignment~\cite{yan2019attention,tel2023alignment}, lightweight architectures~\cite{kong2024safnet}, intrinsic decompositions~\cite{dille2024intrinsic}, and generative merging that hallucinates occluded content~\cite{niu2021hdr,anand2021hdrvideo}. \emph{Spatial} coding instead multiplexes exposures across pixels in a single shot via neutral-density arrays~\cite{nayar2000sve}, coded shutters~\cite{cho2014singleshot}, generalized assorted pixels~\cite{yasuma2010gap}, learned~\cite{nguyen2022sve} or jointly demosaicked~\cite{xu2022jointhdr,chan2023sve22} patterns, and optical randomization~\cite{kakkava2025singleshot}; specialized rigs and dual-readout sensors generalize this idea spatio-temporally~\cite{froehlich14,yue2023hdr}. A distinct family obtains dynamic range without exposure coding: HDR+~\cite{hasinoff2016burst} and its successors~\cite{hasinoff2021bracketing,wadhwa2020livehdr} capture constant-exposure bursts and recover radiance through raw-domain denoising and tone mapping. Extending these ideas to video, alternating-exposure pipelines cycle the shutter between exposures and align neighboring frames via flow or deformable attention~\cite{kalantari2017deep,chen2021coarse,chung2023lanhdr,xu2024hdrflow}, joint interpolation~\cite{khan2022deephs}, or exposure completing~\cite{cui2024exposurecompleting}; a closely related line decodes HDR from naturally occurring auto-exposure variations using self-supervised learning~\cite{banterle2024self} or temporal propagation~\cite{ye2024deep}. None of these methods applies to arbitrary, casually recorded SDR video, but the bracket-and-merge structure they pioneered directly informs our generation-time formulation.

\subsection{Generative Video Models}

\begin{figure}[t]
    \centering
    \includegraphics[width=\columnwidth]{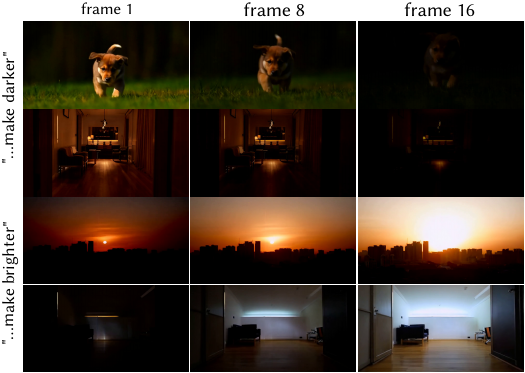}
    \caption{We demonstrate that a pre-trained video model, Wan2.2-I2V-5B,
    has the ability to perform exposure/brightness changes simply by
    prompting it to generate scenes with the keywords ``make darker'' or
    ``make brighter''.}
    \label{fig:gen_understand_exposure}
\end{figure}

Large-scale generative video models~\cite{li2024sora3D,brooks2024video,wan2025wan} model not only motion but implicitly aspects of the image-formation pipeline---exposure, focus, and lens geometry. \emph{Generative Photography}~\cite{generative_photography} makes camera intrinsics an explicit controllable input, enabling applications such as post-capture refocusing~\cite{Tedla2025Refocus}, deblurring~\cite{Tedla2025Blur2Vid}, relighting~\cite{bharadwaj2025genlit}, and cinematic editing~\cite{cinectrl}. Building on this line, we observe that Wan~2.2-I2V-5B~\cite{wan2025wan} responds to exposure-related prompts such as ``make brighter''/``make darker'' without any finetuning (Figure~\ref{fig:gen_understand_exposure}). However, these prompt-driven exposure shifts are not metrically calibrated, not aligned across frames, and do not follow a linear exposure ladder---all properties required for HDR merging. Our finetuning scheme converts this implicit sensitivity into a calibrated, aligned linear exposure-bracket generator.

\begin{figure}[t]
    \centering
    \includegraphics[width=\columnwidth]{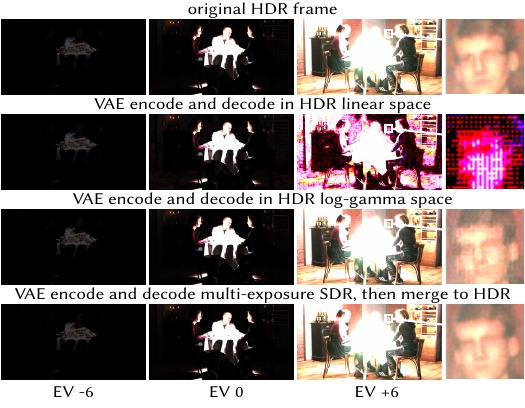}
    \caption{We visualize (i) the exposure bracket of an original HDR frame;
    (ii) direct VAE encoding and decoding in linear space, which causes severe
    artifacts and quality degradation (see $\text{EV} +6$); (iii) VAE encoding
    and decoding in log-gamma space~\cite{yu2026diffhdr}, which produces better
    results but suffers from artifacts in shadow regions (see face crop); and
    (iv) a bracketed approach, encoding and decoding a multi-exposure SDR bracket
    and reconstructing HDR using classical merging~\cite{debevec1997hdr}, which
    results in minimal artifacts. Image from \copyright Stuttgart~\shortcite{froehlich14}.}
    \label{fig:vae_issues}
\end{figure}

\subsection{Learning SDR-to-HDR Conversion}

\noindent\paragraph{Single-image methods.}
With only a single SDR image, recovery is severely ill-posed and clipped highlights and crushed shadows must be plausibly hallucinated. The literature splits into \emph{Inverse Tone Mapping} (ITM)---which targets display-referred HDR (e.g., 8-bit BT.709 to 10-bit BT.2020/PQ) via global tone operators~\cite{akyuz2007hdr,didyk2008enhancement}, CNN highlight reconstruction~\cite{eilertsen2017hdr}, multi-branch networks~\cite{marnerides2018expandnet}, mask guidance~\cite{Santos20SIHDR}, transformers~\cite{wang2025aim}, embedded MLPs~\cite{liu24}, and gain-map representations~\cite{liao25,Canham_2025_ICCV}---and \emph{HDR image inference}, which predicts scene-referred linear radiance. The most relevant inference line, conceptually closest to ours, reformulates the task as \emph{generation-time multi-exposure synthesis} followed by HDR merging. Approaches include pseudo-bracket prediction by regression~\cite{endo2017deep,lee2018deep,zhang2023revisiting} or GANs~\cite{wang2023glowgan}, and diffusion-based bracket consistency~\cite{bemana2024exposure}, which couples multiple SDR denoising trajectories so they agree under re-exposure. LEDiff~\cite{wang2025lediff} finetunes Stable Diffusion to produce latent brackets which are merged in latent-space before decoding; we find this latent-space merging transfers poorly to video (Section~\ref{sec:ablation}). A separate strategy avoids brackets and re-encodes HDR into a pretrained-friendly space: X2HDR~\cite{wu2026x2hdr} adapts an SDR-pretrained diffusion model via a perceptually uniform encoding (PU21/PQ), supporting both text-to-HDR and single-image RAW-to-HDR. Frame-wise application of any of these image-domain priors to video produces severe temporal flickering and exposure drift, since none has temporal modeling.


\noindent\paragraph{Video methods.}
The setting most relevant to ours---HDR recovery from arbitrary, casually recorded SDR video---has only very recently been addressed. Concurrent to our work, DiffHDR~\cite{yu2026diffhdr} (no public code) and LumiVid~\cite{korem2026lumivid} both adapt video foundation models by re-encoding HDR into a single-channel-stretched space (Log-Gamma and LogC3, respectively) and performing radiance inpainting. We instead generate bracketed SDR video that is natively compatible with the video model's latent space. As shown in Figure~\ref{fig:vae_issues}, alternative encoding spaces produce artifacts in the VAE round-trip, whereas our multi-exposure bracketing strategy combined with classical merging~\cite{debevec1997hdr} yields higher-fidelity HDR reconstruction.

\begin{figure*}
    \centering
    \includegraphics[width=1.0\linewidth]{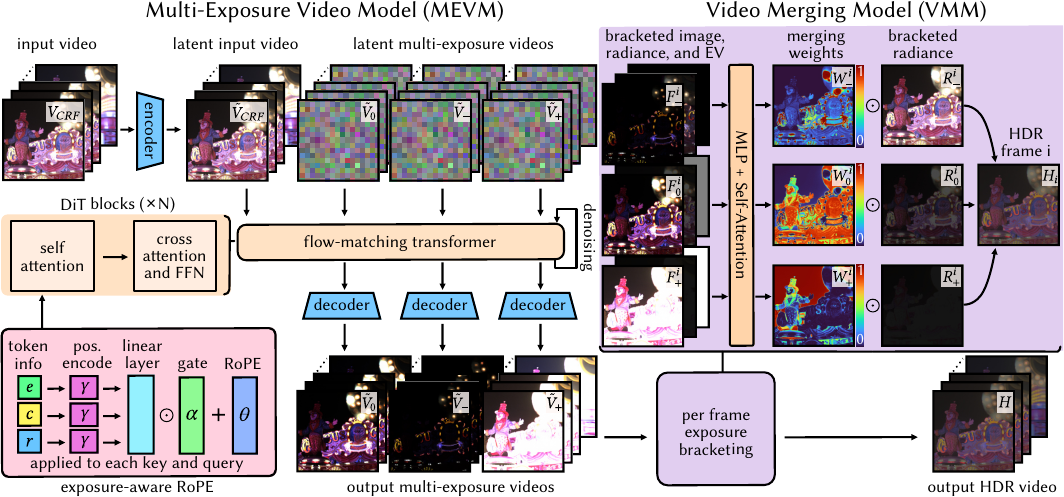}
    \caption{Overview of our HDR video generation pipeline. Our method consists of two stages. \textit{(Left)} The \textbf{Multi-Exposure Video Model (MEVM)} utilizes a flow-matching video transformer (Wan2.2) to generate a linearized multi-exposure video from a single SDR input video $V_{CRF}$. The input video is encoded into a latent $\tilde{V}_{CRF}$ and temporally concatenated with noise-initialized latent frames for the base ($\tilde{V}_0$), under- ($\tilde{V}_-$), and over-exposed ($\tilde{V}_+$) outputs. To enable the model to reason across exposures, we introduce an \textbf{exposure-aware RoPE} that augments the pre-trained rotary positional embedding with a learned offset encoding the exposure index $e$, CRF indicator $c$, and relative frame index $r$. After denoising, the VAE decoder is applied independently to each exposure to recover the output multi-exposure videos. \textit{(Right)} The \textbf{Video Merging Model (VMM)} operates per-frame on the generated exposure bracket. For each frame $i$, per-exposure feature maps $F_k^i$ consisting of the SDR image, estimated radiance, and exposure value are fed into an MLP with self-attention across exposures to predict blending weights $W_k^i$ of the per-exposure radiance estimates $R_k^i$, resulting in the final HDR frame $H_i$. Images from \copyright Stuttgart~\shortcite{froehlich14}.}
    \label{fig:model}
\end{figure*}

\section{Generating Exposure-Bracketed Video}
\label{sec:method}


One of the main challenges in building a scene-referred HDR video generation model is the scarcity of large-scale HDR video data \cite{froehlich14, wang2025lediff}, making direct training of HDR generative models infeasible~\cite{wang2025lediff}. Additionally, finetuning an off-the-shelf SDR model directly on HDR data also fails~\cite{wang2025lediff, korem2026lumivid}, as the latent space of these SDR video models is trained on $8$-bit SDR data, which triggers significant distortion when encoding HDR images as seen in Figure~\ref{fig:vae_issues}. 
We propose breaking this problem into two stages (Figure~\ref{fig:model}): a Multi-Exposure Video Model (MEVM) that accepts an SDR input ($\sdrvid_{CRF}$) and generates a linearized exposure-bracketed video ($\{\sdrvid_0, \sdrvid_-, \sdrvid_+\}$), and a Video Merging Model (VMM) that merges the bracket into the final HDR output ($\hdrvid$) (Section~\ref{sec:merging}):
\begin{equation}
  \sdrvid_{CRF}
  \;\xrightarrow{\;\text{MEVM}\;}\;
  \bigl\{\sdrvid_{0},\,\sdrvid_{-},\,\sdrvid_{+}\bigr\}
  \;\xrightarrow{\;\text{VMM}\;}\;
  \hdrvid.
  \label{eq:pipeline}
\end{equation}

\subsubsection{Preliminary: Video Generation Models}
\label{sec:preliminary_video_models}

Video generation models learn the distribution of natural videos by modeling the probability of a video \vid, optionally conditioned on signals \cond, which may consist of text, images, or videos~\cite{ho2022video}:
\begin{equation}\label{eq:hdr_dist}
    P(\vid \, | \, \cond).
\end{equation}
We can sample from this distribution using a reverse-flow matching process~\cite{wan2025wan}.
To improve memory and efficiency, latent video models~\cite{yang2025cogvideox, wan2025wan} aim to generate a low-resolution latent video \latentsdrvid, where each latent frame represents multiple high-resolution video frames. The final output video \sdrvid can then be recovered by utilizing the pre-trained video decoder. 


\subsection{Multi-Exposure Video Model}

We propose a latent multi-exposure video model (MEVM) that takes latent SDR video with an arbitrary camera response function (CRF) $\latentsdrvid_{CRF}$ as input and generates a latent multi-exposure video $\latentsdrvid_{gen} =\{\latentsdrvid_{0}, \latentsdrvid_{-},\latentsdrvid_{+}\}$ that consists of \textit{linearized} videos of the base exposure $\latentsdrvid_{0}$ (0 EV), a low exposure $\latentsdrvid_{-}$ ($-4$ EV), and a high exposure $\latentsdrvid_{+}$ ($+4$ EV). Formally, MEVM models the conditional probability of generating a multi-exposure video as
\begin{equation}
    P(\latentsdrvid_{gen} | \, \latentsdrvid_{CRF}).
    \label{eq:mevm_dist}
\end{equation}

\subsubsection{Architecture.} We adapt a pre-trained video generation model, Wan2.2-I2V-5B~\cite{wan2025wan}, to support the conditioning in~\eref{eq:mevm_dist}. Specifically, we finetune the 5-billion-parameter video flow-matching transformer~\cite{peebles2023scalable}.

\subsubsection{Latent space encoding.} The pre-trained spatio-temporal variational auto-encoder (VAE) defines the model's latent space and converts videos of size $[\numframes = 4\tilde{F}-3] \times [C=3] \times H\times W$  to size $\tilde{\numframes} \times [\tilde{C} = 48] \times  [\tilde{H} = \frac{H}{16}] \times [\tilde{W} = \frac{W}{16}]$ where $\tilde{F}$ represents the number of latent frames, $\tilde{C}$ is the latent channel dimension, and $\tilde{H} \times \tilde{W}$ is the height and width of the latent feature map. 

\subsubsection{Temporal concatenation.}
\label{sec:concat}
We concatenate the latent CRF SDR video $\latentsdrvid_{CRF}$ and the noise-initialized multi-exposure video along the \emph{temporal} dimension as
\begin{equation}
\textsc{Concat}\big[
\latentsdrvid_{CRF},
\latentsdrvid_{0},
\latentsdrvid_{-},
\latentsdrvid_{+}
\big]
\end{equation}
to generate a latent video with $4\tilde{\numframes}$ frames that is passed to the video flow-matching transformer. We stack exposures along the frame axis (rather than the channel axis) to preserve the pretrained VAE latent dimension and match the model’s expected video input format. After denoising, the VAE decoder is applied separately to each exposure within this latent video to produce a multi-exposure video in pixel space.

\subsubsection{Learnable rotary positional embedding.} Our input, consisting of multiple frames at different exposure levels, forms a set of tokens that must attend not only across space and time, but also across exposure. However, the pre-trained axial rotary positional embedding (RoPE)~\cite{heo2024rotary} within WanVideo~\shortcite{wan2025wan} only represents spatial and temporal indices, and therefore provides no explicit notion of exposure or cross-exposure correspondence. As a result, we find that the pre-trained RoPE is unable to encode this information effectively (see the ablation study in Section~\ref{sec:evaluation}). To address this, we propose a learnable embedding that modulates the pre-trained RoPE, enabling it to encode both exposure level and cross-exposure interactions.

RoPE~\cite{heo2024rotary} encodes spatio-temporal patch position by rotating query and key vectors in the attention block with a position-dependent angle $\theta$ derived from predefined frequencies (see Supplementary Section~S3 for details). To maintain minimal changes to the pretrained model, we introduce a learnable exposure embedding as an offset to the RoPE rotation vector $\theta$ to further embed exposure and relative time information. Specifically, we let each embedding encode three components: the exposure index $e$, the binary CRF indicator $c$, and the relative frame index $r$. We set the exposure index equal to the frame’s exposure value, so frames with identical exposure share the same index. The CRF indicator is a binary value used to distinguish CRF-encoded input frames from linearly generated output frames that share the same exposure level. The relative index groups frames across exposures occurring at the same temporal position. Then, within each DiT's self-attention block, we encode these indices using a standard sinusoidal position encoding $\gamma$~\cite{vaswani2017attention}, concatenate them, and utilize a linear projection layer to offset the pre-trained RoPE embedding, resulting in a modified rotation vector
\begin{equation}
\bar{\theta} =
\theta + \alpha\odot\left(
    \textsc{Linear} \circ \textsc{Concat}\left[\gamma(e),\gamma(c), \gamma(r)\right] \right),
    \label{eq:temporal-encoding}
\end{equation}
\noindent where $\alpha \in \mathbb{R}^{\bar{C}/2}$ is a learnable gate, and $\odot$ is the Hadamard product. We initialize the gate $\alpha$ to 0, so pretraining is unaffected.







\subsubsection{Finetuning.} To finetune the video flow-matching transformer, we sample a linear multi-exposure video from a ground-truth HDR video. We then apply a CRF, noise, and quantization to the middle-exposure video to produce the SDR input video. More details are given in Section~\ref{sec:train_details}. We encode all videos using the pre-trained VAE and concatenate them along the temporal dimension as described in Section~\ref{sec:concat}. We then add noise $\noise \sim \mathcal{N}(0, I)$ to the linear multi-exposure latent frames according to the flow matching process — forming a noisy interpolant $\latentsdrvid_t = t\,\latentsdrvid + (1-t)\,\noise$ at a randomly sampled timestep $t \in [0,1]$ — while keeping the CRF conditioning frames clean. We apply the learnable exposure embedding to the noisy latent frames, patchify them, and feed them to the video flow-matching transformer. We optimize all parameters of the transformer to minimize the flow matching objective
\begin{equation}
    \mathbb{E}_{\latentsdrvid,\,\noise,\,t}\bigl\lVert \hat{v} - (\latentsdrvid - \noise) \bigr\rVert_1,
\end{equation}
i.e., the expected L1 difference between the predicted velocity $\hat{v}$ and the target velocity $\latentsdrvid - \noise$. We use L1 rather than the standard L2 to weight losses more equally across the different exposure brackets, as confirmed in our ablation study (Section~\ref{sec:ablation}).

\section{Merging Exposure-Bracketed Video}
\label{sec:merging}
After generating exposure brackets, we require a method to merge the generated exposure-bracketed video frames into HDR frames. We find classical merging~\cite{debevec1997hdr} reduces perceptual quality, as the VAE introduces small value changes during encoding and decoding that a fixed merging rule cannot correct. We propose a learnable per-pixel video merging model (VMM) that improves the perceptual quality but trades off distortion (see Section~\ref{sec:ablation} for metrics). 

\subsection{Video Merging Model} We perform video merging by learning a lightweight network that operates in a per-frame manner on input exposure-bracketed video. The input to the model is an exposure bracket $\{\sdrvid_{0}^{i}, \sdrvid_{+}^{i}, \sdrvid_{-}^{i}\}$ at a frame $i$ with corresponding exposure values $\{E_{0}^{i}, E_{+}^{i}, E_{-}^{i}\}$, where $E_{k}^{i} \in \mathbb{R}$ denotes the scalar exposure associated with exposure $k$. For each exposure $k \in \{0,+,-\}$, we estimate the scene radiance $\radiance_{k}^{i}$
by normalizing the observed linear pixel value by its exposure:
\begin{equation}
\radiance_{k}^{i} = \frac{\sdrvid_{k}^{i}}{E_{k}^{i}}.
\end{equation}
Then, for each exposure, we construct a 7D feature map
\begin{equation}
F_{k}^{i} =
\left[ \sdrvid_{k}^{i}, \radiance_{k}^{i}, E_{k}^{i} \right],
\end{equation}
concatenating the SDR image, the radiance, and the exposure.

We feed the feature maps $\{F_{k}^{i}\}_{k \in \{0,+,-\}}$ into a
lightweight per-pixel MLP followed by self-attention across exposures, and apply a per-pixel softmax to produce blending weights $\{\weight_{k}^{i}\}$ that sum to one. Finally, we recover the HDR radiance by merging the estimated radiance maps using the predicted weights:
\begin{equation}
\hdrvid^{i} =
\sum_{k \in \{0,+,-\}} \weight_{k}^{i} \odot \radiance_{k}^{i}.
\end{equation}

\subsubsection{Training.} To train the VMM, we sample a linear multi-exposure video from a ground-truth HDR video. We then pass each video through the pre-trained VAE encoder and decoder to simulate distortions introduced by the VAE. The resulting exposure-bracketed video is fed into the VMM and merged into HDR video. This is supervised with an L1 loss in log space~\cite{wang2025lediff}, normalized by the maximum scene radiance $s$ of the ground-truth HDR video:
\begin{equation}
    \mathcal{L} = \bigl\lVert \log(\hdrvid / s + \varepsilon) - \log(\hat{\hdrvid} / s + \varepsilon) \bigr\rVert_1,
\end{equation}
\noindent where $\varepsilon = 10^{-6}$ is a small constant for numerical stability. 

\section{Evaluation}
\label{sec:evaluation}
\subsection{Training Details}
\label{sec:train_details}

\subsubsection{MEVM finetuning} We fine-tune MEVM's flow-matching transformer to generate 17-frame exposure-bracketed clips, and all subsequent inference and evaluation are fixed to this length. For training, we sample 11,336 clips from 26 HDR videos of the Stuttgart dataset \cite{froehlich14}. We train with a batch size of 64 and a learning rate of 0.00003.
Training includes two stages: the first 15 epochs use a lower resolution of $512{\times}768$, and the final 15 epochs use the full resolution of $704{\times}1280$. This requires $\sim$1 week on 8 H100 GPUs.
Each training example consists of a target linear multi-exposure sequence and a corresponding SDR input video, both derived from the same HDR source. At sampling time, MEVM uses 50 steps.

\paragraph{Sampling linear multi-exposure videos}
From each HDR training video, we sample multi-exposure sequences in the linear domain with a fixed 4\,EV spacing between exposures.
The middle (reference) exposure is drawn uniformly at random from a per-scene range $[e_{\min},\, e_{\max}]$, where $e_{\min}$ is the exposure at which 10\% of pixels are clipped to black (0) and $e_{\max}$ is the exposure at which 30\% of pixels are clipped to white (1).

\paragraph{Sampling SDR input videos}
The SDR input to the model is generated from the reference exposure video by applying sensor noise, a random CRF, clipping, and quantization. Details of this procedure are given in Supplementary Section~S4.

\subsubsection{VMM training}
The VMM is trained for 5 epochs on the same Stuttgart training split with the same multi-exposure sampling scheme as the MEVM.
Because the VMM assumes clean, linearized bracketed video, noise and CRF simulation are not applied during VMM training.

\subsection{SDR-to-HDR Video Reconstruction}

We evaluate the performance of our method on 17-frame videos derived from 9 held-out test scenes in Stuttgart~\cite{froehlich14} and 10 scenes from UBC~\cite{azimi2014evaluating}.

\paragraph{Input generation} 
From the ground-truth HDR frames, we generate SDR input videos under two exposure conditions: over- and under-exposure.
For over-exposure, a single fixed exposure is derived from the first frame, with its mean luminance set to 0.70, and this exposure is held constant across all frames.
For under-exposure, the same per-scene fixed scheme is used, but targeting a mean luminance of 0.01 on the first frame. Then, the scaled linear values are clipped to $[0,1]$, gamma-encoded with $\gamma=2.2$, and quantized to 8-bit SDR frames. We additionally explore a third auto-exposed input in Supplementary Section~S2.


\paragraph{Baselines} 
We compared our method against the recent deep-learning-based HDR inverse tone mapping operators: HDRCNN \cite{Eilertsen19StableVideo}, SingleHDR \cite{liu2020single}, LEDiff \cite{wang2025lediff}, X2HDR \cite{wu2026x2hdr}, and LumiVid~\cite{korem2026lumivid}. HDRCNN and LumiVid are video-based methods that explicitly enforce temporal coherence, whereas the remaining methods are applied independently to each frame.
For SingleHDR, LEDiff, X2HDR, and LumiVid, we use the official pre-trained weights.
HDRCNN was retrained following the original protocol~\cite{Eilertsen19StableVideo}, excluding frames from Stuttgart and UBC datasets to prevent data leakage between training and evaluation.
For methods without an explicit linearization stage (HDRCNN and X2HDR), we apply the SingleHDR linearization network \cite{liu2020single} to estimate and invert the CRF. For a given video, the CRF is estimated from the most well-exposed frame selected by the criterion of Mertens~\shortcite{mertens2007exposure}.

Additional quantitative comparisons with HDRTV \cite{Chen211}, DVITMO \cite{xu2019deep}, and ZHDRV \cite{banterle2024self} are provided in Supplementary Section~S2.1.


\paragraph{Metrics}
We follow the metrics protocol of Hanji~\shortcite{hanji2022comparison} and IntrinsicHDR~\cite{dille2024intrinsic}.
Ground-truth frames are preprocessed following Liu et al.~\shortcite{liu2020single} to map values into $[1, 1000]$. For each scene, predictions are then affine-aligned to the calibrated ground truth via least-squares fitting on pixels whose ground-truth luminance lies between the 10th and 90th percentiles, followed by the Hanji et al.~\shortcite{hanji2022comparison} CRF correction.
Following similar work~\cite{wu2026x2hdr, korem2026lumivid}, we report CVVDP~\cite{mantiuk2024colorvideovdp} as our primary quality metric, computed on the CRF-corrected sequences with the following display configuration: 30-inch 4K HDR, BT.709-linear, $L_{\mathrm{peak}}=1000$~cd/m$^2$, $10^6{:}1$ contrast, 10~lux ambient.
We also compute PU-PSNR after the CRF-corrected prediction and ground truth are PU21-encoded~\cite{azimi2021pu21}.
PU-PIQE is computed on the affine-aligned (pre-CRF) prediction as a no-reference quality estimate. To assess temporal stability, we additionally report PU-TF, the VBench Temporal Flickering metric~\cite{huang2023vbench} on the PU-encoded videos.
Following LEDiff~\shortcite{wang2025lediff}, we apply the Drago~\shortcite{drago03} and Mantiuk~\shortcite{mantiuk08} tonemappers directly to the affine-aligned prediction and ground truth (no CRF correction) and compute FID~\cite{heusel2017gans} on the resulting SDR frames (D-FID and M-FID).



\paragraph{Discussion}
Our method achieves the highest or second-highest CVVDP across every exposure condition and both datasets (Table~\ref{tab:stuttgart_ubc_results_short}), while maintaining competitive D-FID and M-FID scores, indicating that our outputs are both perceptually faithful to the ground truth and visually realistic.
Our method outperforms single-image baselines on CVVDP in all but one setting,  confirming the benefit of generating temporally coherent exposure brackets.

Compared to the video baseline LumiVid, our method consistently achieves higher CVVDP; as shown in Figure~\ref{fig:qual_figure_1} and Supp.~Fig.~S3 LumiVid produces prominent high-frequency artifacts due to its encoding strategy, which degrades perceptual quality.
Our method achieves the highest PU-TF among generative methods in three of four settings, with LumiVid narrowly ahead on UBC with under-exposed inputs. Non-generative methods (e.g., SingleHDR) score well on PU-TF because they do not generate motion in clipped regions.
While our method does not achieve the highest PU-PSNR in all settings, we attribute this to the fact that large portions of the output are inpainted, and our generative approach trades distortion for perceptual plausibility~\cite{cohen2023perception}. Please see Supplementary Section~S2 for additional comparisons and analysis.

\begin{table*}[t]
\centering
\caption{Quantitative comparison on the Stuttgart~\shortcite{froehlich14} and UBC~\shortcite{azimi2014evaluating} datasets under different exposure settings.}
\resizebox{\textwidth}{!}{
\begin{tabular}{lllcccccccccccc}
\toprule
  &   &   & \multicolumn{6}{c}{Stuttgart} & \multicolumn{6}{c}{UBC} \\
\cmidrule(lr){4-9} \cmidrule(lr){10-15}
Setting & Type & Method & CVVDP $\uparrow$ & PU-PSNR $\uparrow$ & D-FID $\downarrow$ & M-FID $\downarrow$ & PU-PIQE $\downarrow$ & PU-TF $\uparrow$ & CVVDP $\uparrow$ & PU-PSNR $\uparrow$ & D-FID $\downarrow$ & M-FID $\downarrow$ & PU-PIQE $\downarrow$ & PU-TF $\uparrow$ \\
\midrule
\multirow{6}{*}{Over} & Image & SingleHDR &                      6.00 &  \cellcolor{tabthird}25.32 &  \cellcolor{tabthird}0.95 & \cellcolor{tabsecond}0.70 &                      39.31 &  \cellcolor{tabfirst}0.9959 &                      4.02 &                      19.67 &                      2.01 &                      1.62 &                      46.71 &  \cellcolor{tabfirst}0.9954 \\
 & Image & X2HDR     &                      5.55 &                      24.27 & \cellcolor{tabsecond}0.63 &  \cellcolor{tabfirst}0.63 &                      32.97 &                      0.9873 &                      5.42 &                      21.35 &  \cellcolor{tabthird}0.69 &  \cellcolor{tabthird}0.60 &                      38.12 &                      0.9892 \\
 & Image & LEDiff    &                      5.74 &                      23.62 &                      2.57 &                      3.72 &                      35.26 &                      0.9919 &  \cellcolor{tabthird}5.85 &  \cellcolor{tabthird}22.99 & \cellcolor{tabsecond}0.48 & \cellcolor{tabsecond}0.45 &                      39.75 &  \cellcolor{tabthird}0.9921 \\
 & Video & HDRCNN    &  \cellcolor{tabthird}6.11 & \cellcolor{tabsecond}25.89 &                      1.62 &                      3.51 &  \cellcolor{tabthird}30.54 &                      0.9919 &  \cellcolor{tabfirst}6.64 &  \cellcolor{tabfirst}26.97 &                      1.10 &                      0.88 & \cellcolor{tabsecond}31.29 &                      0.9912 \\
 & Video & LumiVid   & \cellcolor{tabsecond}6.24 &                      25.00 &                      1.08 &                      0.95 &  \cellcolor{tabfirst}26.59 &  \cellcolor{tabthird}0.9922 &                      5.63 &                      21.06 &                      1.25 &                      1.11 &  \cellcolor{tabthird}32.59 &                      0.9919 \\
 & Video & Ours      &  \cellcolor{tabfirst}6.56 &  \cellcolor{tabfirst}27.60 &  \cellcolor{tabfirst}0.61 &  \cellcolor{tabthird}0.79 & \cellcolor{tabsecond}29.79 & \cellcolor{tabsecond}0.9937 & \cellcolor{tabsecond}6.59 & \cellcolor{tabsecond}24.48 &  \cellcolor{tabfirst}0.25 &  \cellcolor{tabfirst}0.20 &  \cellcolor{tabfirst}27.92 & \cellcolor{tabsecond}0.9950 \\
\midrule
\multirow{6}{*}{Under} & Image & SingleHDR &  \cellcolor{tabfirst}7.96 & \cellcolor{tabsecond}29.74 & \cellcolor{tabsecond}1.49 &  \cellcolor{tabthird}1.65 &                      36.72 & \cellcolor{tabsecond}0.9989 &                      6.09 &                      20.73 &                      1.19 &                      1.96 &                      32.72 &  \cellcolor{tabthird}0.9964 \\
 & Image & X2HDR     &                      6.46 &                      26.37 &  \cellcolor{tabthird}1.52 &                      1.73 &                      32.46 &                      0.9892 &  \cellcolor{tabthird}7.09 &  \cellcolor{tabthird}22.27 &  \cellcolor{tabthird}0.82 &  \cellcolor{tabthird}1.16 &  \cellcolor{tabthird}17.67 &                      0.9865 \\
 & Image & LEDiff    &                      6.82 &                      26.09 &                      1.69 &                      2.05 &                      32.51 &                      0.9949 &                      5.19 &                      18.51 &                      0.97 &                      1.74 &                      25.84 &                      0.9958 \\
 & Video & HDRCNN    &                      6.99 &  \cellcolor{tabthird}28.11 &                      2.23 &                      2.62 & \cellcolor{tabsecond}25.62 &                      0.9973 &                      6.17 &                      22.03 &                      1.41 &                      2.17 &                      23.85 &                      0.9959 \\
 & Video & LumiVid   &  \cellcolor{tabthird}7.66 &                      27.09 &                      2.60 & \cellcolor{tabsecond}1.51 &  \cellcolor{tabfirst}24.22 &  \cellcolor{tabthird}0.9988 & \cellcolor{tabsecond}7.91 & \cellcolor{tabsecond}22.74 & \cellcolor{tabsecond}0.54 & \cellcolor{tabsecond}0.62 & \cellcolor{tabsecond}14.36 &  \cellcolor{tabfirst}0.9991 \\
 & Video & Ours      & \cellcolor{tabsecond}7.87 &  \cellcolor{tabfirst}30.71 &  \cellcolor{tabfirst}0.46 &  \cellcolor{tabfirst}0.58 &  \cellcolor{tabthird}27.34 &  \cellcolor{tabfirst}0.9994 &  \cellcolor{tabfirst}8.91 &  \cellcolor{tabfirst}26.84 &  \cellcolor{tabfirst}0.12 &  \cellcolor{tabfirst}0.19 &  \cellcolor{tabfirst}8.38 & \cellcolor{tabsecond}0.9990 \\
\bottomrule
\end{tabular}
}
\label{tab:stuttgart_ubc_results_short}

\end{table*}

\paragraph{User study}
We conducted a perceptual quality study on an HDR monitor following the ITU-R BT.500-13 single stimulus continuous quality evaluation (SSCQE) protocol~\shortcite{itu-bt500}. 
Sixteen observers rated video quality, on a 5-point impairment scale, of outputs from our method and from HDRCNN~\cite{Eilertsen19StableVideo}, LEDiff~\cite{wang2025lediff}, and LumiVid~\cite{korem2026lumivid}.
We selected 5 scenes from the Stuttgart dataset~\cite{froehlich14} and 5 from the UBC dataset~\cite{azimi2014evaluating}, evaluating outputs from both over-exposed and under-exposed input conditions.
Each participant rated 80 videos (4 methods $\times$ 10 scenes $\times$ 2 conditions) in randomized order. 
Additional details are in Supplementary Section~S5. 
Table~\ref{tab:user_study} reports mean SSCQE scores; two-sided Wilcoxon signed-rank tests~\shortcite{woolson2007wilcoxon} (paired on observer $\times$ scene) find our method scores higher than every baseline under both conditions ($p < 0.005$).


\subsubsection{Ablations}
\label{sec:ablation}

Table~\ref{tab:ablation_full} presents an ablation study of three design choices on the Stuttgart test set.
Removing exposure-aware RoPE (row~1) substantially degrades both CVVDP and M-FID for auto-exposed and over-exposed inputs, confirming that an explicit positional signal is necessary to associate tokens across the exposure bracket.
Replacing VMM with classical merging~\cite{debevec1997hdr} (row~2) slightly improves CVVDP but consistently worsens M-FID, indicating a trade-off to our learned merging strategy which targets perceptual quality.
Switching the flow-matching loss from L1 to L2 (row~3) most visibly worsens under-exposure, suggesting L2 biases the model toward generating high-exposure brackets at the expense of lower-exposure brackets.

\begin{table}[t!]
    \centering
    \caption{User study SSCQE scores (ITU-R BT.500-13~\shortcite{itu-bt500}, 1--5 scale).
             We report mean $\pm$ std.~dev.~over 16 observers on 5 Stuttgart and 5 UBC scenes. Under both conditions, the gap between our method and baselines was significant ($p < 0.005$), as determined by the Wilcoxon signed-rank test~\shortcite{woolson2007wilcoxon}. }
    \label{tab:user_study}
    \small
    \setlength{\tabcolsep}{8pt}
    \resizebox{\columnwidth}{!}{%
    \begin{tabular}{lcccc}
        \toprule
        Condition & LEDiff & HDRCNN & LumiVid & Ours \\
        \midrule
        Over
        & $1.94 \pm 0.37$
        & \cellcolor{tabsecond}$2.91 \pm 0.74$
        & \cellcolor{tabthird}$2.37 \pm 0.54$
        & \cellcolor{tabfirst}$3.24 \pm 0.53$ \\

        Under
        & $1.89 \pm 0.32$
        & \cellcolor{tabsecond}$2.76 \pm 0.84$
        & \cellcolor{tabthird}$1.95 \pm 0.44$
        & \cellcolor{tabfirst}$3.50 \pm 0.45$ \\
        \bottomrule
    \end{tabular}}
\end{table}

\begin{table}[t]
    \centering
    \caption{Ablation study on Stuttgart~\shortcite{froehlich14} dataset across three conditions.}
    \label{tab:ablation_full}
    \resizebox{\columnwidth}{!}{%
    \begin{tabular}{ccccccccc}
        \toprule
        \multirow[b]{2}{*}[-1.0ex]{\shortstack{Learn. \\ RoPE}} & \multirow[b]{2}{*}[-1.0ex]{VMM} & \multirow[b]{2}{*}[-1.0ex]{\shortstack{L1 \\ Loss}} & \multicolumn{3}{c}{CVVDP $\uparrow$} & \multicolumn{3}{c}{M-FID $\downarrow$} \\
        \cmidrule(lr){4-6} \cmidrule(lr){7-9}
          &   &   & Auto & Over & Under & Auto & Over & Under \\
        \midrule
         & \checkmark & \checkmark & 7.23 & 6.31 & \cellcolor{tabfirst}7.94 & 0.73 & \cellcolor{tabsecond}0.83 & \cellcolor{tabthird}0.75 \\
        \checkmark &  & \checkmark & \cellcolor{tabfirst}7.72 & \cellcolor{tabfirst}6.62 & \cellcolor{tabsecond}7.87 & \cellcolor{tabsecond}0.56 & 0.89 & \cellcolor{tabsecond}0.66 \\
        \checkmark & \checkmark &  & \cellcolor{tabthird}7.43 & \cellcolor{tabthird}6.54 & 7.83 & \cellcolor{tabthird}0.56 & \cellcolor{tabthird}0.83 & 0.99 \\
        \checkmark & \checkmark & \checkmark & \cellcolor{tabsecond}7.67 & \cellcolor{tabsecond}6.56 & \cellcolor{tabthird}7.87 & \cellcolor{tabfirst}0.40 & \cellcolor{tabfirst}0.79 & \cellcolor{tabfirst}0.58 \\
        \bottomrule
    \end{tabular}
    }
\end{table}

\begin{table}[t]
\centering
\caption{Latent-space merging (LEDiff) compared with pixel-space merging via a CNN (DeepHDR) and our lightweight pixel-space merger (VMM) on Stuttgart (Auto). Memory is reported per frame, except for LEDiff, where it is reported per latent frame.}
\label{tab:ltm_stuttgart_results}
\resizebox{\linewidth}{!}{
\begin{tabular}{lcccc}

\toprule
Merger & Params (K) $\downarrow$ & Mem.~(MB) $\downarrow$ & CVVDP $\uparrow$ & M-FID $\downarrow$ \\
\midrule
LEDiff & \cellcolor{tabsecond}157 & \cellcolor{tabfirst}35 & \cellcolor{tabthird}5.02 & \cellcolor{tabthird}5.72\\
DeepHDR  &  \cellcolor{tabthird}384 &  \cellcolor{tabthird}1377 &  \cellcolor{tabsecond}7.18 &  \cellcolor{tabsecond}0.69 \\
VMM (ours)  &  \cellcolor{tabfirst}5 &  \cellcolor{tabsecond}440 &  \cellcolor{tabfirst}7.67 &  \cellcolor{tabfirst}0.40 \\
\bottomrule
\end{tabular}
}
\end{table}

Table~\ref{tab:ltm_stuttgart_results} compares VMM against two alternative merging strategies. LEDiff~\cite{wang2025lediff} performs merging in the VAE's latent space with an MLP; naively adapting this approach to our video model fails to converge to usable results.
DeepHDR~\cite{kalantari2017deep} instead merges in pixel space using a CNN, trained on the same data as our VMM.
VMM outperforms both LEDiff and DeepHDR across all quality metrics while using fewer parameters than either, and less memory than DeepHDR. Qualitatively, DeepHDR produces color shifts and over-smoothed outputs relative to our per-pixel VMM (see Supplementary Figure~S2).



\subsection{Towards Fully Generative HDR Video}
Our pipeline can be chained with an off-the-shelf SDR video model to form a fully generative HDR video pipeline. We prompt a text-to-video model~\cite{wan2025wan} with exposure-related keywords such as ``very bright'', ``sunny day'', ``dark'', and ``no lights'' to produce an input SDR video that can be lifted to HDR with our model. Figure~\ref{fig:qual_figure_1}(b) and Figure~\ref{fig:qual_figure_2}(a, top-left) show examples of this application. Across these qualitatively different generative scenarios, our method produces physically plausible dynamic-range expansion in both highlight and shadow regions while preserving the temporal coherence of the upstream generative video.


\subsection{In-the-Wild Videos}
\begin{figure*}[]
    \centering
    \includegraphics[width=0.87\textwidth]{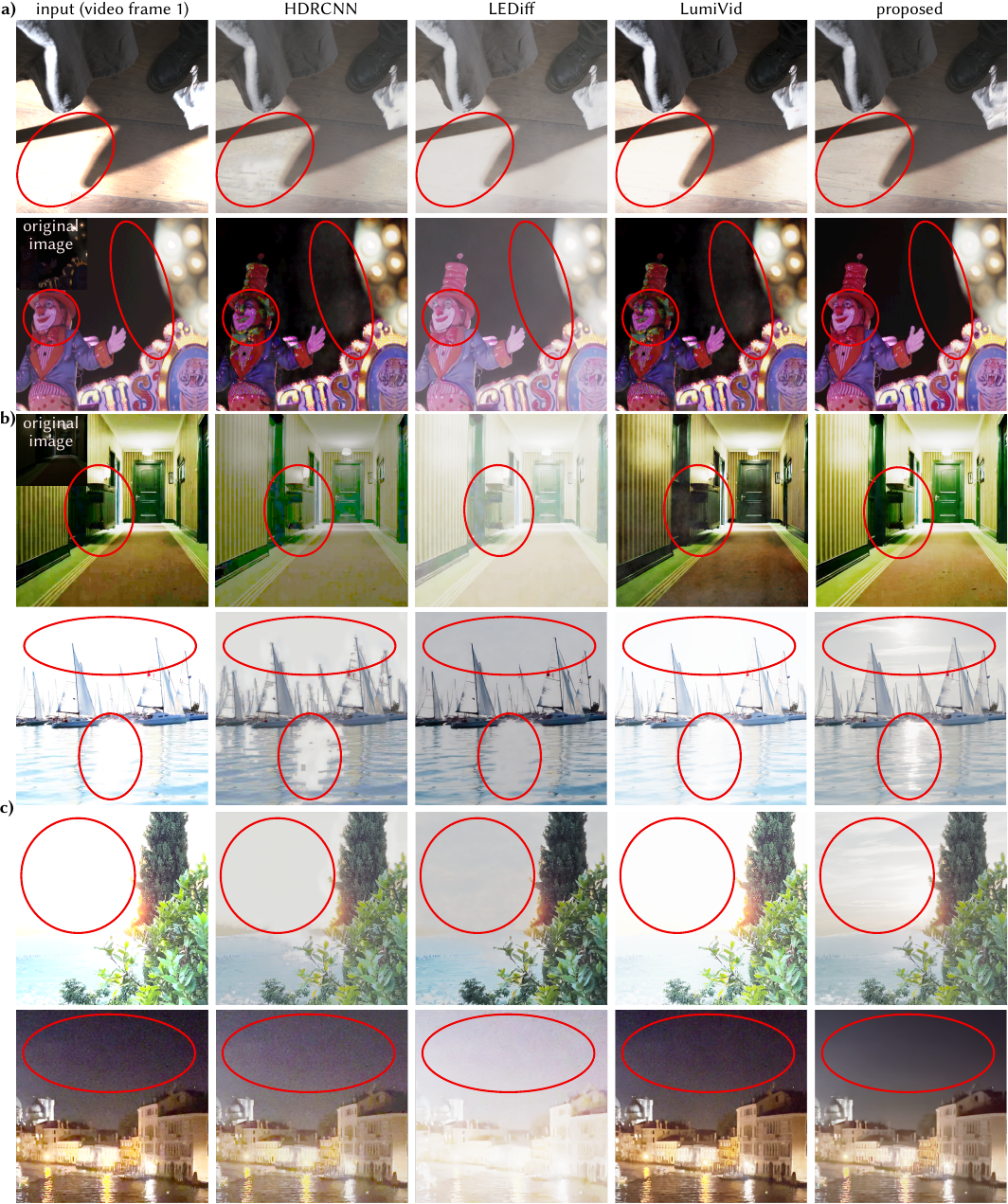}
\caption{Qualitative SDR-to-HDR comparison of HDRCNN~\cite{Eilertsen19StableVideo}, LEDiff~\cite{wang2025lediff}, LumiVid~\cite{korem2026lumivid}, and our method. HDR outputs are visualized with Reinhard~\shortcite{reinhard02} tonemapping. (a)~Two frames from the Stuttgart~\shortcite{froehlich14} and UBC~\shortcite{azimi2014evaluating} HDR video benchmarks, illustrating an \emph{over-exposed} scenario (top), where our method generates a realistic floor while in baselines it remains clipped, and a challenging \emph{under-exposed} carousel scene (bottom), where baseline methods generate green artifacts on the clown's face and hallucinations in the sky.
    (b)~Two frames generated from text-to-SDR-video prompts and then lifted to HDR by each method: \emph{``A hallway in an old hotel with most lights off''} (top, severely under-exposed) and \emph{``A row of sailboats rocking in a harbor on a sunny day''} (bottom, over-exposed). In the hallway scene, we observe our method has fewer quantization artifacts on the radiator and door and for the sailboats scene, our method generates realistic sky and water highlights, whereas they remain clipped in baseline outputs.
    (c)~Two in-the-wild examples — a sunlit Valpolicella hillside framed by cypress/olive trees and a night-time Venice canal scene — where only our method reconstructs a realistic sky, generating realistic highlights in one and denoising shadows in the other. See the \supp{supplementary webpage} for video comparisons (use ``Side-by-side'' mode). }
    \label{fig:qual_figure_1}
  \end{figure*}

We apply our method to in-the-wild SDR videos to demonstrate generalization across a wide variety of scenes and cameras. We show examples in Figure~\ref{fig:qual_figure_1}(c) and Figure~\ref{fig:qual_figure_2}(a, last three images), spanning a range of challenging scenes: a sunlit hillside in Valpolicella; nighttime in Venice; a glowing campfire; and BMX riders in a sun-drenched warehouse. Across all scenes, our method recovers plausible highlight and shadow detail without temporal flickering. See the \supp{supplementary webpage} for more in-the-wild results.

\subsection{Cinematic Moments in HDR}
\begin{figure*}[]
    \centering
    \includegraphics[width=0.88\textwidth]{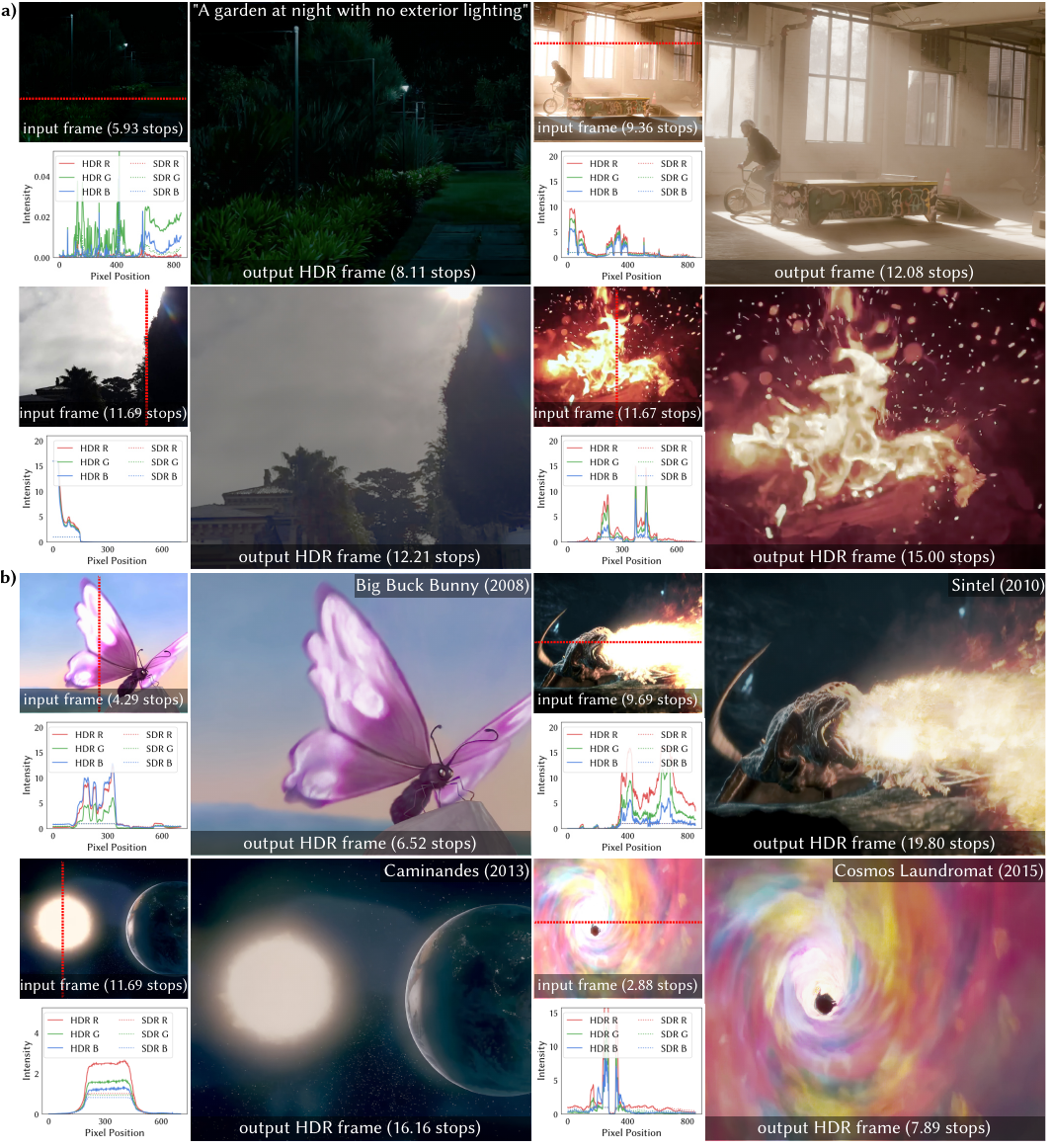}
    \caption{Applications of our pipeline in the wild. For each example, we show the SDR input (top-left) and our HDR result (main panel; Reinhard~\shortcite{reinhard02} tonemapping), annotated with their respective dynamic ranges (in stops, defined as $\log_2$ of the 99th-to-1st percentile luminance ratio; computed on linearized sRGB for the SDR input, and directly on the linear HDR output). We include intensity scan lines (along the red-dashed line) where scanline radiance is relative and unitless, anchored by setting the SDR input's linear range (EV 0) to $[0,1]$; with our $\pm4$EV spacing, EV$+4$ resolves shadows (0–$1/16$), EV0 the midtones (0–1), and EV$-4$ the highlights (0–16), and merging combines these into a single radiance map spanning 0–16. (a)~\emph{Fully generative HDR video and in-the-wild examples.} A text-to-SDR-video model~\cite{wan2025wan} is prompted with ``no exterior lighting'' to produce a dark SDR input, which our model lifts to HDR (top left). We also show three additional in-the-wild examples (BMX riders and fireplace, \copyright~coverr.co). (b)~\emph{Cinematic Moments in HDR.} Our pipeline re-grades cinema frames: A purple butterfly with an overexposed wing poised to take flight (\textit{Big Buck Bunny}, 2008); a dragon breathing fire at a girl, who has tracked it to its mountain lair (\textit{Sintel}, 2010); a blown-out sun next to the dark Earth (\textit{Caminandes}, 2013); and Franck the sheep getting sucked into a colorful storm with a glowing vortex (\textit{Cosmos Laundromat}, 2015). Our method recovers dynamic range in both highlights and shadows while minimizing artifacts from noise or compression. We note that the Sintel scene's dynamic range is expanded by 10 stops, exceeding the $\pm4$EV span, because our generated brackets---unlike a camera's---are not quantized, so their floating-point representation can encode more of the scene's dynamic range. Please see \supp{supplementary webpage} for videos. Films provided by \copyright Blender Foundation under CC BY 3.0.}
    \label{fig:qual_figure_2}
  \end{figure*}

We demonstrate our method on cinema footage. Figure~\ref{fig:qual_figure_2}(b) shows four examples: a purple butterfly with a sunlit wing preparing to take flight (\textit{Big Buck Bunny}, 2008); a dragon breathing fire at Sintel, who has tracked it to its mountain lair (\textit{Sintel}, 2010); a glowing yellow sun next to the dark Earth (\textit{Caminandes}, 2013); and Franck the sheep being pulled into a colorful storm with a glowing vortex (\textit{Cosmos Laundromat}, 2015). Despite the domain gap introduced by film grading and compression artifacts, our pipeline produces temporally coherent HDR outputs that enhance detail in both highlight and shadow regions without amplifying compression or noise. 

\section{Conclusion}
\label{sec:conclusion}

\begin{figure}[t]
    \centering
    \includegraphics[width=0.95\columnwidth]{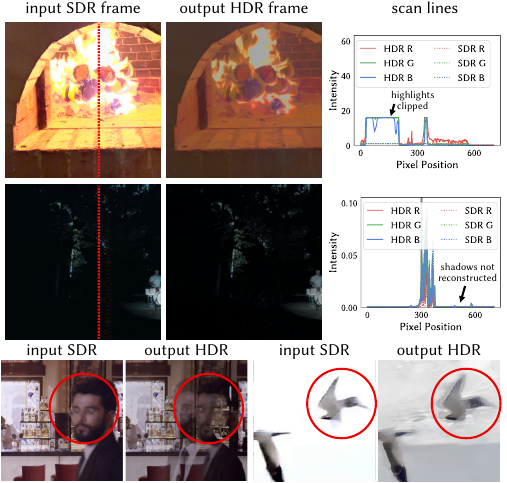}
    \caption{\textit{(Top)} When highlights are extremely bright, the generated $-4\,\text{EV}$ bracket is insufficient to recover the full dynamic range. \textit{(Middle)} For very dark scenes, generating one bracket alone cannot recover shadow detail. \textit{(Bottom)} We observed ghosting/halo artifacts in some HDR outputs due to misalignment in generated brackets. This is more visible in scenes with fast motion or frame cuts. See \supp{supplementary webpage} for videos. }
    \label{fig:limitations}
  \end{figure}

We have shown that a reframing of HDR generation---generating bracketed video at multiple exposures---provides a robust path for converting SDR video to HDR.
Remarkably, our model, finetuned on \emph{fewer than 30 HDR videos}, synthesizes plausible HDR content even on challenging in-the-wild footage, demonstrating the strength of the video model's prior for exposure-conditioned generation.
Our method maintains high performance across benchmarks, generalizes to a wide variety of in-the-wild scenes, including cinematic footage, and can be paired with standard SDR video generation models to create a full generative HDR pipeline.

\paragraph{Limitations and future work.} Our design choice to generate exposure brackets comes at the cost of reduced video length and increased computational cost (Supplementary Table~S3).  
Additionally, our fixed $\pm4$EV three-bracket exposure strategy cannot always recover the full dynamic range of high-contrast scenes, leaving some outputs with residual clipping after generation; we tried wider spacings and even an adaptive spacing strategy but found they introduced color artifacts (see Supplementary Section~S2.2). We also explored diffusion forcing~\cite{chen2025diffusion} to address longer video generation but observed compounding artifacts when conditioning on generated content---a problem of active interest~\cite{po2025baggerbackwardsaggregationmitigating}.

We also observed our model outputs occasionally include halo or ghosting artifacts due to misalignment in generated brackets as well as compression artifacts from the VAE. These artifacts are more visible in scenes with fast motion or frame cuts.
We visualize in Figure~\ref{fig:limitations} and provide further analysis in Supplementary Section~S2.6. 

We note our SDR training generation pipeline omits local tone mapping, color grading, and compression artifacts common in real-world footage. Additionally, our method generates relative radiance, hence per-scene alignment is required for computing metrics.


Finally, our work belongs to a growing class of methods that remap computational photography problems as tasks for video generation models, exploiting the strong camera-physics priors these models learn from internet-scale data~\cite{generative_photography}. Our findings open the door to end-to-end system design, in which auto-exposure algorithms could be designed knowing that a strong generative prior will reconstruct HDR in post-processing.
HDR-native generative models are a present need---and our results suggest repurposing video models for generation-time bracketing offers a practical path forward, without requiring large-scale HDR training data.

\begin{acks}
We thank Siddhu Tedla for help with evaluations. We are grateful to Shuchen Du and Shuo Lei for allocating us their GPUs. We thank the users from the user study for their valuable feedback.  DBL and KNK acknowledge support of NSERC under the RGPIN program.
\end{acks}


\clearpage
\setcounter{page}{1}
\setcounter{section}{0}
\setcounter{figure}{0}
\setcounter{table}{0}
\renewcommand{\thesection}{S\arabic{section}}
\renewcommand{\thesubsection}{S\arabic{section}.\arabic{subsection}}
\renewcommand{\thesubsubsection}{S\arabic{section}.\arabic{subsection}.\arabic{subsubsection}}
\renewcommand{\thefigure}{S\arabic{figure}}
\renewcommand{\thetable}{S\arabic{table}}

\suppmaketitle{Supplementary Material: Generating HDR Video from SDR Video}

\begin{table*}[b]
\centering
\caption{Quantitative comparison on the Stuttgart~\shortcite{froehlich14} and UBC~\shortcite{azimi2014evaluating} datasets under different exposure settings.}
\resizebox{\textwidth}{!}{
\begin{tabular}{lllcccccccccccc}
\toprule
  &   &   & \multicolumn{6}{c}{Stuttgart} & \multicolumn{6}{c}{UBC} \\
\cmidrule(lr){4-9} \cmidrule(lr){10-15}
Setting & Type & Method & CVVDP $\uparrow$ & PU-PSNR $\uparrow$ & D-FID $\downarrow$ & M-FID $\downarrow$ & PU-PIQE $\downarrow$ & PU-TF $\uparrow$ & CVVDP $\uparrow$ & PU-PSNR $\uparrow$ & D-FID $\downarrow$ & M-FID $\downarrow$ & PU-PIQE $\downarrow$ & PU-TF $\uparrow$ \\
\midrule
\multirow{9}{*}{Auto} & Image & SingleHDR &  \cellcolor{tabthird}7.07 &  \cellcolor{tabthird}27.62 &  \cellcolor{tabthird}0.96 &  \cellcolor{tabthird}0.68 &                      36.07 &  \cellcolor{tabfirst}0.9967 &  \cellcolor{tabthird}8.15 &  \cellcolor{tabfirst}29.80 &                      0.74 &                      0.50 &                      36.01 & \cellcolor{tabsecond}0.9979 \\
 & Image & HDRTV     &                      6.40 &                      26.60 &                      1.00 &                      1.28 &                      41.04 & \cellcolor{tabsecond}0.9966 &                      7.77 &                      28.19 &                      0.77 &                      0.98 &                      41.48 &  \cellcolor{tabfirst}0.9983 \\
 & Image & X2HDR     &                      6.07 &                      24.92 &  \cellcolor{tabthird}0.96 &                      0.96 &                      32.25 &                      0.9892 &                      7.97 &                      26.55 &                      0.27 &  \cellcolor{tabfirst}0.20 &                      29.50 &                      0.9908 \\
 & Image & LEDiff    &                      6.05 &                      25.16 &                      2.22 &                      3.27 &                      35.69 &                      0.9928 &                      7.29 &                      26.34 &                      0.54 &                      0.56 &                      34.83 &                      0.9934 \\
 & Video & DVITMO    &                      6.36 &                      26.21 &                      1.12 &                      1.48 &  \cellcolor{tabfirst}23.20 &                      0.9949 &                      7.91 &  \cellcolor{tabthird}28.75 & \cellcolor{tabsecond}0.17 &  \cellcolor{tabfirst}0.20 &  \cellcolor{tabfirst}23.38 &                      0.9958 \\
 & Video & ZHDRV     &                      6.47 &                      26.68 &                      2.25 &                      2.18 & \cellcolor{tabsecond}25.63 &                      0.9935 &                      5.23 &                      20.77 &                      1.71 &                      2.38 &                      27.84 &                      0.9936 \\
 & Video & HDRCNN    &                      6.71 &                      27.32 &                      1.42 &                      2.90 &                      29.18 &                      0.9937 & \cellcolor{tabsecond}8.26 & \cellcolor{tabsecond}29.74 &  \cellcolor{tabfirst}0.15 &  \cellcolor{tabthird}0.23 &  \cellcolor{tabthird}26.35 &                      0.9940 \\
 & Video & LumiVid   & \cellcolor{tabsecond}7.53 & \cellcolor{tabsecond}28.54 &  \cellcolor{tabfirst}0.41 &  \cellcolor{tabfirst}0.37 &  \cellcolor{tabthird}26.62 &                      0.9948 &                      7.92 &                      25.15 &                      0.84 &                      0.73 &                      30.15 &                      0.9958 \\
 & Video & Ours      &  \cellcolor{tabfirst}7.67 &  \cellcolor{tabfirst}29.76 & \cellcolor{tabsecond}0.52 & \cellcolor{tabsecond}0.40 &                      29.35 &  \cellcolor{tabthird}0.9955 &  \cellcolor{tabfirst}8.33 &                      27.36 &  \cellcolor{tabthird}0.24 & \cellcolor{tabsecond}0.22 & \cellcolor{tabsecond}24.97 &  \cellcolor{tabthird}0.9966 \\
\midrule
\multirow{9}{*}{Over} & Image & SingleHDR &                      6.00 &                      25.32 &  \cellcolor{tabthird}0.95 & \cellcolor{tabsecond}0.70 &                      39.31 &  \cellcolor{tabfirst}0.9959 &                      4.02 &                      19.67 &                      2.01 &                      1.62 &                      46.71 & \cellcolor{tabsecond}0.9954 \\
 & Image & HDRTV     &                      5.80 &                      25.55 &                      1.07 &                      1.02 &                      42.36 & \cellcolor{tabsecond}0.9951 &                      5.56 &  \cellcolor{tabthird}23.39 &                      0.86 &                      0.93 &                      35.60 &  \cellcolor{tabfirst}0.9958 \\
 & Image & X2HDR     &                      5.55 &                      24.27 & \cellcolor{tabsecond}0.63 &  \cellcolor{tabfirst}0.63 &                      32.97 &                      0.9873 &                      5.42 &                      21.35 &                      0.69 &                      0.60 &                      38.12 &                      0.9892 \\
 & Image & LEDiff    &                      5.74 &                      23.62 &                      2.57 &                      3.72 &                      35.26 &                      0.9919 &  \cellcolor{tabthird}5.85 &                      22.99 & \cellcolor{tabsecond}0.48 & \cellcolor{tabsecond}0.45 &                      39.75 &                      0.9921 \\
 & Video & DVITMO    &                      5.54 &                      25.32 &                      1.47 &                      2.12 &  \cellcolor{tabfirst}20.87 &  \cellcolor{tabthird}0.9937 &                      5.17 &                      22.69 &  \cellcolor{tabthird}0.64 &  \cellcolor{tabthird}0.57 &  \cellcolor{tabfirst}19.75 &                      0.9935 \\
 & Video & ZHDRV     &                      5.80 & \cellcolor{tabsecond}26.10 &                      2.28 &                      2.47 &                      30.42 &                      0.9933 &                      5.47 &                      22.86 &                      1.52 &                      1.64 &  \cellcolor{tabthird}29.34 &                      0.9923 \\
 & Video & HDRCNN    &  \cellcolor{tabthird}6.11 &  \cellcolor{tabthird}25.89 &                      1.62 &                      3.51 &                      30.54 &                      0.9919 &  \cellcolor{tabfirst}6.64 &  \cellcolor{tabfirst}26.97 &                      1.10 &                      0.88 &                      31.29 &                      0.9912 \\
 & Video & LumiVid   & \cellcolor{tabsecond}6.24 &                      25.00 &                      1.08 &                      0.95 & \cellcolor{tabsecond}26.59 &                      0.9922 &                      5.63 &                      21.06 &                      1.25 &                      1.11 &                      32.59 &                      0.9919 \\
 & Video & Ours      &  \cellcolor{tabfirst}6.56 &  \cellcolor{tabfirst}27.60 &  \cellcolor{tabfirst}0.61 &  \cellcolor{tabthird}0.79 &  \cellcolor{tabthird}29.79 &  \cellcolor{tabthird}0.9937 & \cellcolor{tabsecond}6.59 & \cellcolor{tabsecond}24.48 &  \cellcolor{tabfirst}0.25 &  \cellcolor{tabfirst}0.20 & \cellcolor{tabsecond}27.92 &  \cellcolor{tabthird}0.9950 \\
\midrule
\multirow{9}{*}{Under} & Image & SingleHDR &  \cellcolor{tabfirst}7.96 & \cellcolor{tabsecond}29.74 & \cellcolor{tabsecond}1.49 &  \cellcolor{tabthird}1.65 &                      36.72 &  \cellcolor{tabthird}0.9989 &                      6.09 &                      20.73 &                      1.19 &                      1.96 &                      32.72 &                      0.9964 \\
 & Image & HDRTV     &                      5.50 &                      27.02 &                      1.81 &                      1.75 &                      50.33 &  \cellcolor{tabfirst}0.9997 &                      5.10 &                      19.48 &                      2.01 &                      2.32 &                      48.94 &  \cellcolor{tabfirst}0.9999 \\
 & Image & X2HDR     &                      6.46 &                      26.37 &  \cellcolor{tabthird}1.52 &                      1.73 &                      32.46 &                      0.9892 &  \cellcolor{tabthird}7.09 &  \cellcolor{tabthird}22.27 &  \cellcolor{tabthird}0.82 &  \cellcolor{tabthird}1.16 &                      17.67 &                      0.9865 \\
 & Image & LEDiff    &                      6.82 &                      26.09 &                      1.69 &                      2.05 &                      32.51 &                      0.9949 &                      5.19 &                      18.51 &                      0.97 &                      1.74 &                      25.84 &                      0.9958 \\
 & Video & DVITMO    &                      7.01 &                      27.54 &                      2.31 &                      2.10 &  \cellcolor{tabfirst}16.99 &                      0.9977 &                      5.91 &                      19.82 &                      2.47 &                      3.52 & \cellcolor{tabsecond}13.64 &                      0.9954 \\
 & Video & ZHDRV     &                      6.39 &                      26.03 &                      3.70 &                      2.97 & \cellcolor{tabsecond}20.99 &                      0.9967 &                      2.30 &                      16.57 &                      2.39 &                      2.00 &                      31.21 &                      0.9945 \\
 & Video & HDRCNN    &                      6.99 &  \cellcolor{tabthird}28.11 &                      2.23 &                      2.62 &                      25.62 &                      0.9973 &                      6.17 &                      22.03 &                      1.41 &                      2.17 &                      23.85 &                      0.9959 \\
 & Video & LumiVid   &  \cellcolor{tabthird}7.66 &                      27.09 &                      2.60 & \cellcolor{tabsecond}1.51 &  \cellcolor{tabthird}24.22 &                      0.9988 & \cellcolor{tabsecond}7.91 & \cellcolor{tabsecond}22.74 & \cellcolor{tabsecond}0.54 & \cellcolor{tabsecond}0.62 &  \cellcolor{tabthird}14.36 & \cellcolor{tabsecond}0.9991 \\
 & Video & Ours      & \cellcolor{tabsecond}7.87 &  \cellcolor{tabfirst}30.71 &  \cellcolor{tabfirst}0.46 &  \cellcolor{tabfirst}0.58 &                      27.34 & \cellcolor{tabsecond}0.9994 &  \cellcolor{tabfirst}8.91 &  \cellcolor{tabfirst}26.84 &  \cellcolor{tabfirst}0.12 &  \cellcolor{tabfirst}0.19 &  \cellcolor{tabfirst}8.38 &  \cellcolor{tabthird}0.9990 \\
\bottomrule
\end{tabular}
}
\label{tab:stuttgart_ubc_results_full}
\end{table*}

The supplementary material is organized as follows: the supplementary webpage (Section~\ref{sec:supp_webpage}), additional results and discussion (Section~\ref{sec:additional_results}), architecture details (Section~\ref{sec:architecture_details}), training details (Section~\ref{sec:supp_train_details}), and user study details (Section~\ref{sec:supp_user_study}).

\section{Supplementary Webpage}
\label{sec:supp_webpage}
We provide an interactive supplementary webpage for viewing results. Open \texttt{webpage/index.html} in Chrome (113 or later) to launch the viewer; the page has been tested on a Mac HDR display. Upon opening, you will be prompted to select the \texttt{videos/} folder included with this supplement. The viewer supports a variety of tonemappers for visualization and presents side-by-side comparisons against the strongest baselines.

The webpage is organized into four sections. In-the-wild results show our method applied to real-world SDR video sourced from the internet, with no ground truth available. Text-to-HDR results demonstrate a fully generative pipeline in which text-to-video outputs serve as SDR inputs, producing HDR content entirely from text. Cinematic moments present frames sampled from open-source films spanning a wide range of tonal regimes. Finally, limitations collects failure cases for transparency and to motivate future work.

\section{Additional Results and Discussion}
\label{sec:additional_results}

\subsection{Quantitative Results}
\label{sec:quantitative_results}

Table~\ref{tab:stuttgart_ubc_results_full} reports the complete quantitative comparison, including additional baselines omitted from the main text due to their significantly lower performance: HDRTV \cite{Chen211}, DVITMO \cite{xu2019deep}, and ZHDRV \cite{banterle2024self}. HDRTV is evaluated using the official pre-trained weights. DVITMO is retrained following the original implementation because pre-trained weights are not publicly available. ZHDRV is self-supervised and is therefore optimized independently for each input scene. For methods without an explicit linearization stage (DVITMO, HDRTV, and ZHDRV), we apply the SingleHDR linearization network~\cite{liu2020single} to estimate and invert the CRF. For a given video, the CRF is estimated from the most well-exposed frame selected by the criterion of Mertens~\shortcite{mertens2007exposure}.

Table~\ref{tab:stuttgart_ubc_results_full} also explores another ``auto-exposure'' input case, in which we produce inputs with a per-frame exposure so that the mean scene luminance maps to 0.25 after scaling. We then temporally smooth the exposure values using a 3-frame uniform moving average, which helps suppress abrupt flicker~\cite{tedla2023examining}.

\subsection{Exposure Bracket Ablation}
\label{sec:supp_ablations}

We ablate our fixed three-exposure bracket with $\pm4$EV spacing against tighter ($\pm2$EV) and wider ($\pm6$EV, $\pm8$EV) fixed spacings, as well as an adaptive strategy. Rather than offsetting the bracket by a fixed EV amount from the input's own exposure, the adaptive strategy selects a base exposure so the scene's highest radiance maps to 0.9 in the $+4$EV bracket; the predicted exposures are therefore determined by scene content rather than relative to the input exposure.

\begin{figure}[t]
  \centering
  \vspace{-10pt}
  \includegraphics[width=0.5\textwidth]{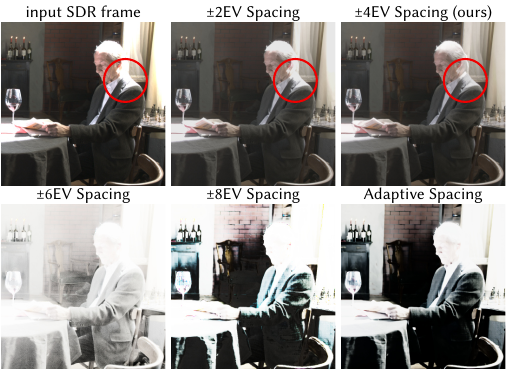}
  \caption{Effect of different bracket-spacing strategies on an over-exposed scene. HDR outputs are visualized with Reinhard~\shortcite{reinhard02} tonemapping. The narrower $\pm2$EV spacing leaves highlights on the subject's neck and shoulder clipped. Wider spacings ($\pm6$EV, $\pm8$EV) produce merging and color-shift artifacts that worsen with spacing; the adaptive, scene-dependent spacing similarly introduces a visible color cast. Image from \copyright Stuttgart~\shortcite{froehlich14}.}
  \label{fig:bracket_ablation}
  \vspace{-10pt}
\end{figure}

Table~\ref{tab:bracket_ablation} shows that our $\pm4$EV spacing achieves the best CVVDP, PU-PSNR, M-FID, and D-FID. As shown in Figure~\ref{fig:bracket_ablation}, wider spacings ($\pm6$EV, $\pm8$EV) produce large color shifts and artifacts after merging, while the narrower $\pm2$EV spacing does not adequately resolve the highlight and shadow regions clipped in the input. The adaptive strategy performs worst overall, as the model struggles to generate brackets that are evenly spaced before merging. This indicates a naive approach is not sufficient, but we believe implementing adaptive spacing is a promising direction for future work, as it would allow the model to select brackets dependent on the scene's content.

\begin{table}[h]
\centering
\caption{Ablation of the exposure-bracket sampling strategy on Stuttgart (Auto). We compare our fixed $\pm4$EV spacing against tighter ($\pm2$EV) and wider ($\pm6$EV, $\pm8$EV) fixed spacings, and an adaptive spacing strategy.}
\label{tab:bracket_ablation}
\resizebox{\linewidth}{!}{
\begin{tabular}{lccccc}
\toprule
Strategy & CVVDP $\uparrow$ & PU-PSNR $\uparrow$ & M-FID $\downarrow$ & D-FID $\downarrow$ & PU-PIQE $\downarrow$ \\
\midrule
$\pm2$EV & \cellcolor{tabsecond}6.45 & \cellcolor{tabsecond}26.70 & \cellcolor{tabthird}1.71 & \cellcolor{tabthird}0.91 & 32.24 \\
$\pm4$EV (ours) & \cellcolor{tabfirst}7.67 & \cellcolor{tabfirst}29.76 & \cellcolor{tabfirst}0.40 & \cellcolor{tabfirst}0.52 & \cellcolor{tabsecond}29.35 \\
$\pm6$EV & \cellcolor{tabthird}5.95 & \cellcolor{tabthird}25.39 & \cellcolor{tabsecond}0.69 & \cellcolor{tabsecond}0.67 & \cellcolor{tabthird}30.71 \\
$\pm8$EV & 4.72 & 23.50 & 3.29 & 3.18 & \cellcolor{tabfirst}27.24 \\
Adaptive & 4.83 & 23.48 & 4.23 & 3.82 & 47.16 \\
\bottomrule
\end{tabular}
}
\end{table}

\begin{table}[t]
  \centering
  \caption{Inference time and peak GPU memory for converting a 17-frame video at $1280{\times}704$. Image methods are applied frame-by-frame.}
  \label{tab:inference_cost}
  \small
  \setlength{\tabcolsep}{6pt}
  \begin{tabular}{llcc}
  \toprule
  Method & Type & Inference Time (s) $\downarrow$ & Memory (GB) $\downarrow$ \\
  \midrule
  SingleHDR & Image & \cellcolor{tabsecond}1.6 & \cellcolor{tabsecond}2.4 \\
  X2HDR & Image & 205.9 & 37.7 \\
  LEDiff & Image & 734.9 & \cellcolor{tabthird}17.0 \\
  HDRCNN & Video & \cellcolor{tabfirst}1.3 & \cellcolor{tabfirst}1.5 \\
  LumiVid & Video & \cellcolor{tabthird}29.5 & 22.1 \\
  Ours & Video & 52.1 & 36.5 \\
  \bottomrule
  \end{tabular}
  \end{table}

\subsection{Comparing Merging Methods}
\label{sec:supp_merging_methods}

\begin{figure}[t]
  \centering
  \includegraphics[width=\columnwidth]{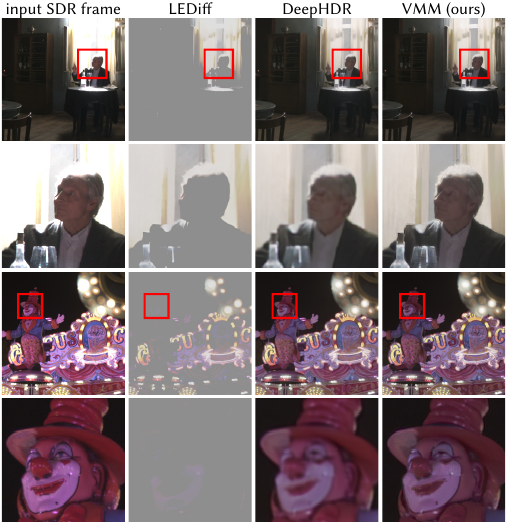}
  \caption{Qualitative comparison of different merging strategies on two Stuttgart scenes, tonemapped for display with Reinhard~\shortcite{reinhard02} tonemapping. LEDiff's~\cite{wang2025lediff} latent-space merging produces washed-out colors. DeepHDR~\cite{kalantari2017deep} merges in pixel space and better preserves color, but its outputs are noticeably over-smoothed, losing fine detail. Our VMM preserves both colors and fine detail. Images from \copyright Stuttgart~\shortcite{froehlich14}.}
  \label{fig:deephdr_compare}
\end{figure}

As discussed in the main paper (Table~\ref{tab:ltm_stuttgart_results}), we compare our per-pixel VMM against two alternative merging strategies: LEDiff's latent-space merging and DeepHDR's~\cite{kalantari2017deep} pixel-space CNN merging. Figure~\ref{fig:deephdr_compare} shows qualitative results.

\subsection{Additional Comparisons with LumiVid}
\begin{figure*}[]
  \centering
  \includegraphics[width=0.9\textwidth]{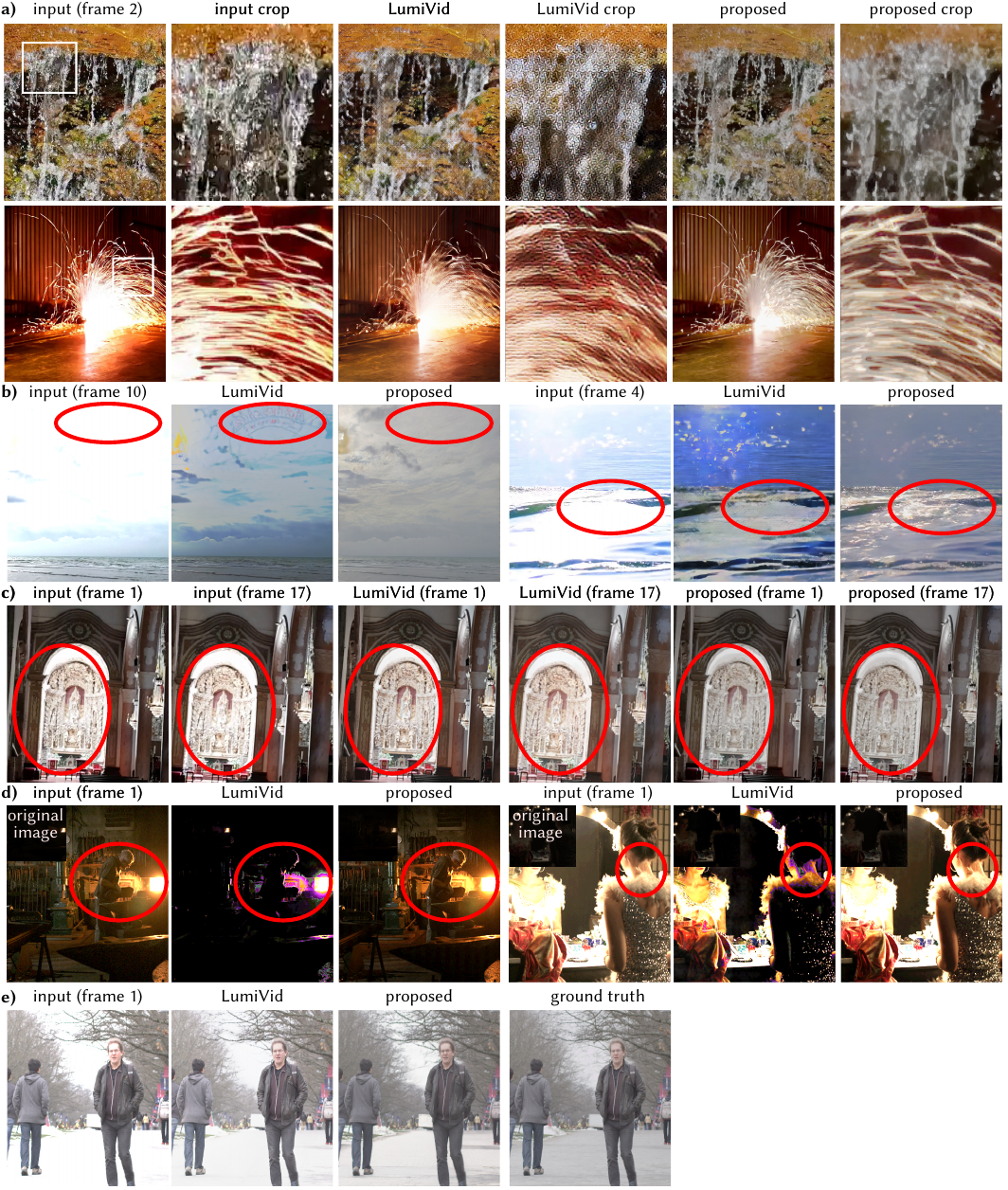}
  \caption{Qualitative comparison between our method (left) and LumiVid~\cite{korem2026lumivid} (right) across five challenging conditions.
  \textit{(a) High-frequency motion:} our method preserves spatial structure in a waterfall scene; LumiVid exhibits irregular moiré-like distortions.
  \textit{(b) Scene fidelity:} our method reconstructs scene content faithfully; LumiVid produces output that diverges significantly from the input.
  \textit{(c) Temporal consistency:} our method maintains stable color across frames; the chapel interior color shifts noticeably in LumiVid's output.
  \textit{(d) Low-light performance:} our method recovers shadow detail; LumiVid introduces prominent pink chrominance artifacts in the same dark regions.
  \textit{(e) Scene-linear output:} our explicit linearization stage correctly inverts the CRF; LumiVid's output retains residual tonemapping in this scene, indicating an incomplete CRF inversion. See \supp{supplementary webpage} for videos.}
  \label{fig:lumivid_compare}
\end{figure*}

Figure~\ref{fig:lumivid_compare} provides additional qualitative comparisons against LumiVid~\cite{korem2026lumivid}, the most closely related baseline. The comparisons highlight several conditions where the two methods differ in output quality. We believe a potential cause of these differences is that our method generates exposure-bracketed SDR sequences that reside \textit{natively} within the pre-trained latent space of the underlying video model, whereas LumiVid encodes HDR in LogC3 — a space that is not inherently compatible with the SDR model it finetunes.

\paragraph{High-frequency motion (a).}
In scenes with fast-moving, high-frequency patterns, such as waterfalls, our method preserves spatial structure, whereas LumiVid exhibits irregular patterns in its output. We hypothesize that this difference arises from our approach operating natively in SDR space, whereas LumiVid must generalize its LogC3 HDR encoding to rapidly changing texture statistics.

\paragraph{Scene fidelity (b).}
Our method reconstructs scene content faithfully under challenging conditions. In contrast, LumiVid can produce outputs with hallucinations that diverge significantly from the input scene, suggesting that its finetuned diffusion model occasionally decouples from the conditioning signal.

\paragraph{Temporal consistency (c).}
In some scenes, where our method produces consistent output, LumiVid outputs videos with perceptible color shifts over time.

\paragraph{Low-light performance (d).}
In dark regions, our method recovers shadow detail from the generated high-exposure bracket. LumiVid introduces prominent pink chrominance artifacts in the same conditions, consistent with unreliable color estimation in low-signal regimes.

\paragraph{Scene-linear output (e).}
Our pipeline learns to output linear SDR frames before merging brackets into HDR. In some scenes, LumiVid's output retains residual tonemapping, indicating the CRF was not fully inverted and the output is not truly scene-linear.

That said, our design decisions have other limitations, which we discuss in Section~\ref{sec:supp_limitations}.

\subsection{Luminance Distribution}
\begin{figure}[h!]
  \centering
  \includegraphics[width=\columnwidth]{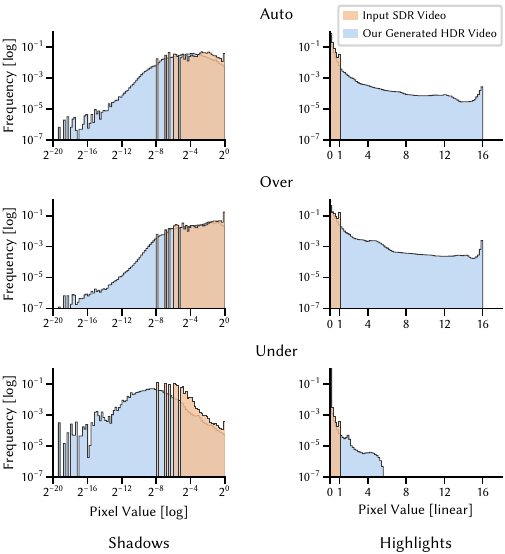}
  \caption{We visualize the luminance histograms of the input SDR and output HDR videos, with statistics averaged across all three scenarios (Auto, Over, and Under) on both datasets (Stuttgart and UBC). To make both ends of the range legible, we plot the same histogram twice---zoomed on shadows (left) and on highlights (right). Frequencies are visualized as percentages on a log scale. We observe that our method consistently expands both the shadow and highlight detail in all scenarios. }
  \label{fig:luminance_contrast_distribution}
\end{figure}

Figure~\ref{fig:luminance_contrast_distribution} compares luminance histograms of the input SDR and output HDR videos across Auto, Over, and Under exposures on Stuttgart and UBC. Our method consistently expands both shadow and highlight content relative to the SDR input; however, highlights clip near a relative luminance of $16$ due to our fixed bracketing strategy. We discuss this limitation in Section~\ref{sec:supp_limitations}.

\subsection{Limitations}
\label{sec:supp_limitations}

\begin{figure}[h!]
  \centering
  \includegraphics[width=\columnwidth]{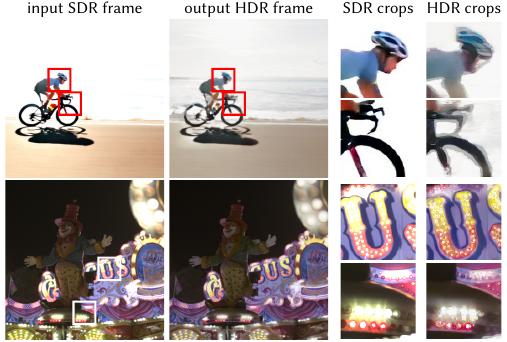}
  \caption{\textit{(Top)} With fast motion and high-frequency details such as the structure of the bike's wheels, the generated brackets can be misaligned, causing ghosting and halo artifacts. \textit{(Bottom)} Latent compression by the video model's VAE introduces visible artifacts; the crops highlight misalignment and spatio-temporal artifacts in our generated HDR. Image from \copyright Stuttgart~\shortcite{froehlich14}.}
  \label{fig:limitations_supp}
\end{figure}

As discussed in the main paper, our method exhibits four main failure modes, illustrated in  Figure~\ref{fig:limitations} (main paper) and Supplementary Figure~\ref{fig:limitations_supp}. First, because we generate a fixed -4EV bracket, scene highlights bright enough to saturate even this darkened exposure cannot be recovered. Second, for extremely dark inputs whose SDR regions are already heavily quantized or clipped, the generated high-exposure bracket provides little usable signal—or only noisy signal—for reconstructing shadow detail. Both cases share a common root cause: some pixels are not well-exposed in any of the three generated brackets. Third, in scenes with fast motion or frame cuts, the generated brackets can be misaligned, causing ghosting and halo artifacts. We believe this is because the VAE performs latent interpolation while decoding, which is imperfect and fundamentally unable to handle scenes with frame cuts. Fourth, the video model's VAE introduces latent compression artifacts, which manifest as a loss of high-frequency detail.

Additionally, our bracketed generation yields higher quality and temporal coherence than other generative SoTA methods, but uses roughly $3{\times}$ the tokens of LumiVid and therefore higher memory (Table~\ref{tab:inference_cost}).
Long-video generation is a limitation for all methods that rely on Wan-type models~\cite{chen2025diffusion}, not just ours; per-frame methods such as LEDiff avoid this constraint at the cost of worse temporal coherence.
LEDiff, LumiVid, and our method thus occupy different points in a memory--quality design space: our focus has been maximum quality (Table~\ref{tab:stuttgart_ubc_results_full}), though hybrid approaches may be possible.

\section{Architecture Details}
\label{sec:architecture_details}

\subsection{WanVideo} We utilize WanVideo~\cite{wan2025wan} as our base video flow-matching transformer. The video flow-matching transformer (DiT) patchifies latent frames into $N$ tokens of dimension $\hat{C}$, which are passed through a series of DiT blocks~\cite{wan2025wan}. Each DiT block contains a self-attention module with multiple attention heads, each processing tokens with $\bar{C}$ channels. The input tokens to each head are used to compute queries $\queries\in \mathbb{R}^{N\times\bar{C}}$ and keys $\keys\in \mathbb{R}^{N\times\bar{C}}$, which are encoded with an axial rotary positional embedding (RoPE)~\cite{heo2024rotary} to produce rotated queries $\queries'$ and keys $\keys'$ used to compute an attention matrix $A=\queries'\keys'^{\top}$.

RoPE is implemented by converting each query and key vector $\queries_n, \keys_m \in \mathbb{R}^{\bar{C}}$ into complex vectors $\bar{\queries}_n, \bar{\keys}_m \in \mathbb{C}^{\bar{C}/2}$ where $\bar{\queries}_{(n,k)}=\queries_{(n,2k)} + i\queries_{(n,2k+1)}$ and $\bar{\keys}_{(m,k)}=\keys_{(m,2k)} + i\keys_{(m,2k+1)}$, taking even entries as real components and odd entries as imaginary components. Each complex vector is then rotated by applying
\begin{equation}
\bar{\queries}'_n=Re[\bar{\queries}_n \circ e^{i\theta(p_n)}], \quad \bar{\keys}'_m=Re[\bar{\keys}_m \circ e^{i\theta(p_m)}]
\end{equation}
where $p = (p_f, p_h, p_w)$ is the token's spatio-temporal coordinates, $\theta(p) \in \mathbb{R}^{\bar{C}/2}$ is a position-dependent rotation vector, $Re[\cdot]$ is the real part of a complex number, and $\circ$ is the Hadamard product. The rotation vector encodes spatio-temporal coordinates and is defined as
\begin{equation}
\theta(p)
=
\big[
\underbrace{\phi(p_f,\,\bar{C}/6)}_{\text{frame}},\,
\underbrace{\phi(p_h,\,\bar{C}/6)}_{\text{height}},\,
\underbrace{\phi(p_w,\,\bar{C}/6)}_{\text{width}}
\big]^{\!\top}
\end{equation}
where $\phi(f,d)\in\mathbb{R}^{d}$ is defined elementwise as
\begin{equation}
\phi(f,d)_t = 10000^{-t f / d}, \qquad t=0,\ldots,d-1,
\end{equation}
and $[\cdot,\cdot,\cdot]^{\top}$ denotes vector concatenation.

\paragraph{Exposure-aware RoPE} We modify this RoPE embedding to encode exposure level and cross-exposure correspondence as described in Section~\ref{sec:method}.

\subsection{Video Merging Model}

The VMM operates per-frame and per-pixel on the three exposure brackets. For each frame $i$ and exposure $k \in \{0,+,-\}$, the 7D feature map $F_k^i = [\sdrvid_k^i,\, \radiance_k^i,\, E_k^i]$ is the input to the network, where $\sdrvid_k^i \in \mathbb{R}^3$ is the SDR image, $\radiance_k^i \in \mathbb{R}^3$ is the estimated scene radiance, and $E_k^i \in \mathbb{R}$ is the scalar exposure value.

\paragraph{Per-pixel MLP} Each 7D feature $F_k^i$ at every pixel is independently projected to a $d$-dimensional embedding via a two-layer MLP:
\begin{equation}
    \mathbf{z}_k^i = \textsc{Linear}_{d_{\mathrm{h}} \to d}\bigl(\textsc{GELU}\bigl(\textsc{Linear}_{7 \to d_{\mathrm{h}}}(F_k^i)\bigr)\bigr),
\end{equation}
where we use hidden dimension $d_{\mathrm{h}} = 128$ and output dimension $d = 64$.

\paragraph{Exposure self-attention.} The three embeddings $\{\mathbf{z}_0^i, \mathbf{z}_+^i, \mathbf{z}_-^i\}$ at each pixel are treated as a sequence of length 3 and passed through a multi-head self-attention block~\cite{vaswani2017attention} with $h{=}4$ heads. To preserve the per-pixel nature of the network, all spatial pixels are batched into the batch dimension so that attention is computed \emph{only across exposures}, with no spatial interaction:
\begin{equation}
    \mathbf{z}_k^i \leftarrow \mathbf{z}_k^i + \textsc{MHA}\bigl(\textsc{LN}(\mathbf{z}_0^i), \textsc{LN}(\mathbf{z}_+^i), \textsc{LN}(\mathbf{z}_-^i)\bigr),
\end{equation}
where $\textsc{LN}$ denotes layer normalization applied before attention (pre-norm residual).

\paragraph{Weight prediction.} A final linear head projects each embedding to a scalar logit, and a softmax over exposures produces the blending weights:
\begin{equation}
    \weight_k^i = \textsc{Softmax}_k\!\bigl(\textsc{Linear}_{d \to 1}(\textsc{LN}(\mathbf{z}_k^i))\bigr).
\end{equation}
The HDR frame is then recovered as the weighted sum of per-exposure radiance estimates as described in Section~\ref{sec:merging} of the main paper.

\section{Training Details}
\label{sec:supp_train_details}
During training, we simulate realistic videos by applying sensor noise, a random CRF, clipping, and quantization to the clean reference videos. 

  \begin{figure*}[]
    \centering
    \includegraphics[width=0.9\textwidth]{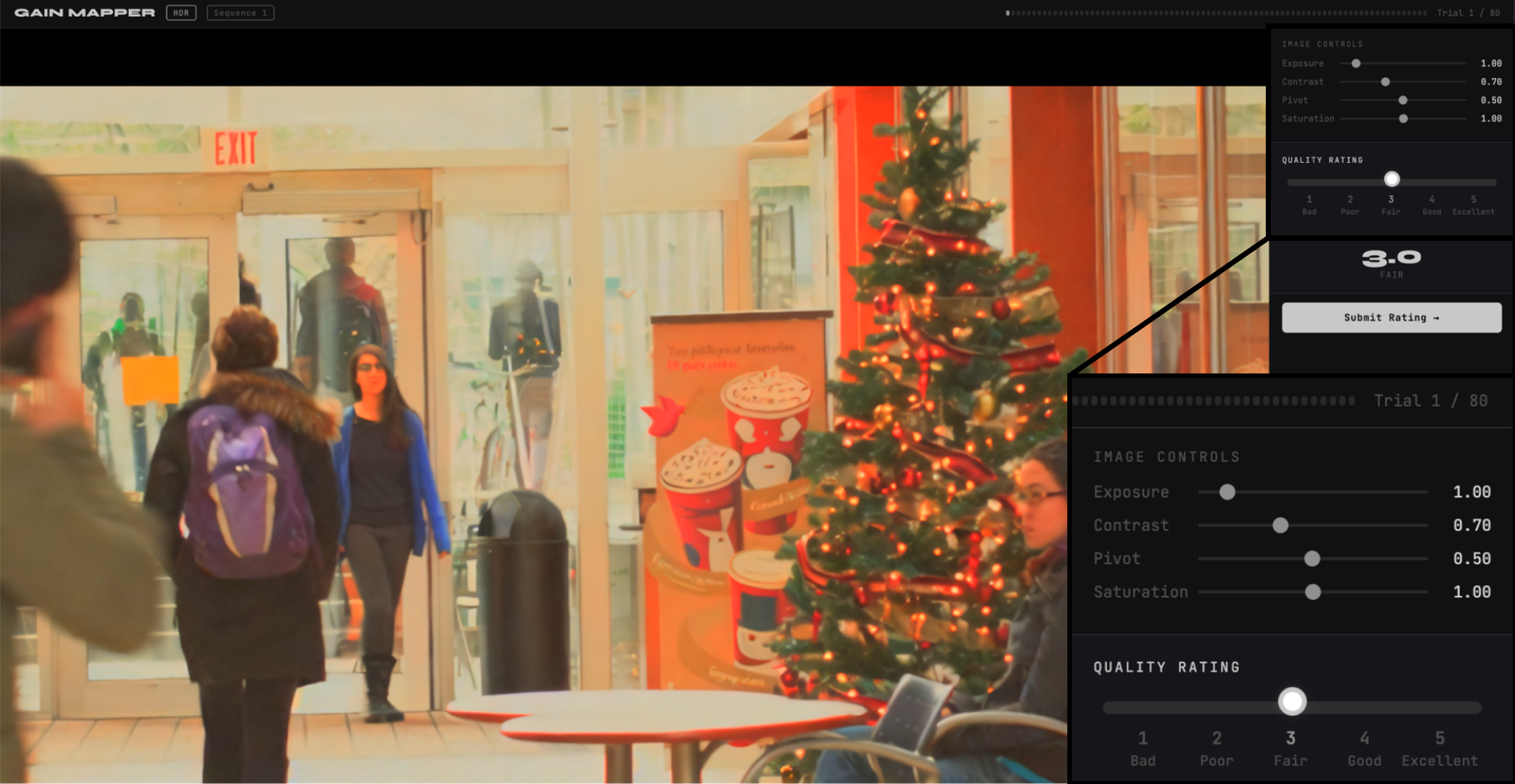}
    \caption{User Study Interface. Tonemapping and rating sliders are shown at an enlarged size. Image from \copyright UBC~\shortcite{azimi2014evaluating}.}

    \label{fig:ui}
  \end{figure*}

\paragraph{Sensor noise.} To simulate sensor noise, we apply Poisson-Gaussian noise~\cite{plotz2017benchmarking} to each training sequence in the linear domain before the CRF is applied.
Given a frame $i$ with pixel intensity $x^i$, per-pixel noise $n(x^i)$ is drawn from a heteroscedastic model as
\begin{equation}
    n(x^i) \sim \sqrt{\sigma_s^2\, x^i + \sigma_r^2}\ \epsilon^i,
\end{equation}
where $\sigma_s \sim \mathcal{U}(0,\, 0.05)$ and $\sigma_r \sim \mathcal{U}(0,\, 0.02)$ are shot and read noise parameters, and $\epsilon^i$ represents temporally-correlated Gaussian noise. 
Specifically, we model this noise with a first-order autoregressive process \cite{box2015time},
\begin{equation}
    \epsilon^i = \rho\,\epsilon^{i-1} + \sqrt{1-\rho^2}\,u^i, \quad u^i \sim \mathcal{N}(0,\,\mathbf{I}),
\end{equation}
where $\rho=0.5$ for all frames except the first frame, for which $\rho=0$. 

\paragraph{CRF sampling.}
To approximate the diversity of real camera response curves, we follow Eilertsen et al.~\shortcite{eilertsen2017hdr} and apply a parametric CRF of the form
\begin{equation}
    f(H_{i,c}) = (1 + \sigma)\,\frac{H_{i,c}^{n}}{H_{i,c}^{n} + \sigma}
    \label{eq:crf}
\end{equation}
to each training sequence, where $n \sim \mathcal{N}(0.9, 0.1)$ and $\sigma \sim \mathcal{N}(0.6, 0.1)$.

\paragraph{Clipping and quantization.} Finally, we clip the CRF-encoded values to $[0,1]$ and quantize them to 8-bit SDR frames.

\section{User Study Details}
\label{sec:supp_user_study}
We conducted a user study following the ITU-R BT.500-13 single stimulus continuous quality evaluation (SSCQE) protocol~\shortcite{itu-bt500}.
Sixteen observers rated video quality on a 5-point impairment scale (1 = Bad, 5 = Excellent), preceded by two practice trials.
Videos were displayed on a MacBook built-in Liquid Retina XDR display at 1,600 $\text{cd}/\text{m}^2$ peak brightness via a custom WebGPU shader equipped with exposure and tonemapping controls.
Tonemapping was implemented via a Naka-Rushton equation \cite{banic18} applied to exposure-adjusted scene-linear luminance.
Tonemapped luminance is then applied to scene linear RGB channels via the color correction method of Mantiuk et al. \cite{mantiuk09}, leaving the saturation parameter open for observers to adjust.
Via the experimental interface shown in Figure \ref{fig:ui}, observers could modulate controls for exposure, contrast and pivot, or the image intensity where the contrast parameter is most impactful.
This enabled users to display the image at their desired exposure and contrast and evaluate the image as editing source material.
Videos were played on loop so that observers had as much time as they liked to adjust and assess them.

\bibliographystyle{ACM-Reference-Format}
\bibliography{main}

\end{document}